\documentclass{article} %
\usepackage{air_opd_preprint,times}

\usepackage{amsmath,amsfonts,bm}

\def\eqref#1{equation~\ref{#1}}

\def\1{\bm{1}}

\DeclareMathAlphabet{\mathsfit}{\encodingdefault}{\sfdefault}{m}{sl}
\SetMathAlphabet{\mathsfit}{bold}{\encodingdefault}{\sfdefault}{bx}{n}

\newcommand{\KL}{D_{\mathrm{KL}}}

\usepackage[hypertexnames=false]{hyperref}
\hypersetup{hidelinks,pdftitle={Learning from Evolving Errors: Adaptive Iterative Repair for On-Policy Distillation},pdfauthor={Rui Li, Liyang He, Zheng Zhang, Zhenya Huang, Linbo Zhu, Qi Liu},pdfsubject={Preprint}}
\usepackage{url}
\usepackage{booktabs}
\usepackage{multirow}
\usepackage{graphicx}
\usepackage{wrapfig}
\usepackage{needspace}
\usepackage{amsmath}
\usepackage{amssymb}
\usepackage{dsfont}
\usepackage{algorithm}
\usepackage{algorithmic}
\usepackage{array}
\usepackage{tabularx}
\usepackage{microtype}
\usepackage{float}   %
\usepackage{caption}
\usepackage[most]{tcolorbox}

\newcolumntype{Y}{>{\centering\arraybackslash}X}
\newcolumntype{L}{>{\raggedright\arraybackslash}X}

\definecolor{promptboxback}{gray}{0.955}
\definecolor{anchorhl}{rgb}{1.0,0.92,0.75}
\newtcolorbox{promptbox}[2][]{
  enhanced,
  breakable,
  width=\linewidth,
  colback=promptboxback,
  colframe=black,
  colbacktitle=black,
  coltitle=white,
  boxrule=0.7pt,
  arc=2mm,
  outer arc=2mm,
  left=3mm,
  right=3mm,
  top=1.6mm,
  bottom=1.8mm,
  toptitle=1.0mm,
  bottomtitle=1.0mm,
  before skip=6pt,
  after skip=8pt,
  fonttitle=\small\bfseries,
  fontupper=\small,
  before upper={\setlength{\parskip}{4pt}\setlength{\parindent}{0pt}},
  title={#2},
  #1
}

\newcommand{\promptrule}{%
  \par\nobreak\vspace{1pt}%
  \noindent\textcolor{black!25}{\rule{\linewidth}{0.4pt}}%
  \par\vspace{-1pt}%
}

\newcommand{\methodname}{\textsc{Air-OPD}}

\title{Learning from Evolving Errors:\\
Adaptive Iterative Repair for On-Policy Distillation}

\author{Rui Li$^{1}$ \enspace Liyang He$^{1}$ \enspace Zheng Zhang$^{2}$ \enspace
Zhenya Huang$^{1}$ \enspace Linbo Zhu$^{1}$ \enspace Qi Liu$^{1}$\\[0.5em]
{\normalfont\normalsize $^{1}$University of Science and Technology of China}\\[0.15em]
{\normalfont\normalsize $^{2}$Nanyang Technological University}\\[0.35em]
{\normalfont\small\texttt{ruili2000@mail.ustc.edu.cn}}}

\begin{document}

\maketitle

\begin{abstract}
On-policy self-distillation (OPSD) supplies dense token-level feedback on
trajectories sampled from the student's own policy, a richer training
signal than the outcome-level rewards of reinforcement learning. This
feedback comes from a teacher conditioned on a full reference solution
unavailable to the student. The reference solution specifies the target
but not how to move from the student's current error toward it,
creating a \emph{solution-conditioned shortcut risk}.
We introduce \methodname{}, an adaptive iterative repair framework for
on-policy distillation that provides error-to-repair supervision. Given a failed response, a
guidance generator synthesizes
repair guidance for the current error. The student samples an
on-policy retry with this guidance. If the retry remains incorrect, the generator
produces new repair guidance for the newly observed error. At each round, a
fixed teacher receives the guidance as privileged context and supervises the
student on an error-aligned region of its latest failed response.
Outcome-aware stage weighting favors early repair stages and credits
stages whose immediate retry passes verification.
We train \methodname{} on the DAPO-Math-17K dataset
and evaluate on AIME24, AIME25, and HMMT25, alongside
out-of-distribution tests on MMLU-Pro and GPQA. We examine two guidance
sources, self-guidance from the current student policy and external guidance from a
larger model. For both Qwen3-4B and Qwen3-8B, \methodname{} attains the best
mathematical-reasoning averages, improving over the strongest baseline by up
to 3.6 points, while preserving base-model performance on the
out-of-distribution benchmarks.

\end{abstract}

\section{Introduction}
\label{sec:introduction}

Reinforcement learning with verifiable rewards (RLVR) is now a standard
approach to improving language-model reasoning in post-training
\citep{shao2024deepseekmath,yu2025dapo}. However, typical RLVR algorithms such
as GRPO \citep{guo2025deepseekr1,shao2024deepseekmath} rely on outcome rewards
to compute response-level advantages for policy optimization. Such
response-level supervision does not distinguish fine-grained token-level
contributions within a trajectory. On-policy distillation (OPD) complements
this outcome signal with dense token-level supervision on the student's own
trajectories \citep{agarwal2024onpolicy,lu2026opd}. On-policy
self-distillation (OPSD) further removes the need for a separate teacher
model \citep{zhao2026selfdistilled}. Through in-context learning, the model
serves as its own teacher, conditioned on a reference solution as
privileged context that the student never observes.

Learning with privileged information uses training-only signals to guide
the learner \citep{vapnik2009new,lopezpaz2016unifying}. In
reference-conditioned OPSD, a successful attempt already satisfies the
verifier, and further conditioning on a static reference may steer the
student toward the reference's solution path. Privileged supervision is thus
most informative on failed attempts, and recent work treats successful and
failed trajectories asymmetrically \citep{hou2026uniopd,zheng2026scope}. For failed
trajectories, a reference solution identifies the correct destination but
does not specify the transition from the student's current error to a
valid continuation. When the teacher
remains conditioned on that same full solution, the student may learn to
imitate privileged demonstrations rather than acquire generalizable
error-correction mechanisms. We refer to this as the
\emph{solution-conditioned shortcut risk}.

This leads to our central research question:

\begin{quote}
\emph{How can privileged supervision turn a failed on-policy trajectory into a
sequence of learnable error-to-repair transitions?}
\end{quote}

To answer this question, we introduce \methodname{}. Inspired by instructional
scaffolding \citep{wood1976role}, \methodname{} performs multi-round adaptive
repair that replaces a static reference-conditioned target with repair
guidance synthesized for the student's latest error. Given a failed response, a privileged
guidance generator uses the reference solution to synthesize
repair guidance as scaffolding for the identified error, and the student samples an on-policy
retry with this guidance. If the retry remains incorrect, the generator synthesizes fresh
repair guidance targeting the newly observed error. At every round, the guidance serves as teacher-only
privileged context for supervising the student on its latest failed
response. \methodname{} supervises only an error-aligned repair region of
that response and excludes tokens that the repair guidance does not
target, which would otherwise add noise. Each round thus contributes one stage loss. To combine these losses
across rounds, we design outcome-aware stage weighting, which favors
earlier repair stages and rewards stages whose retry passes verification.

We train on mathematical problems from DAPO-Math-17K
\citep{yu2025dapo}, pairing each problem with a reference solution
generated by GPT-5.5, and evaluate on AIME24,
AIME25, and HMMT25. We additionally test transfer beyond this mathematical
training distribution on MMLU-Pro and GPQA, and compare two guidance
sources, self-guidance from the current student policy and external guidance from
a larger model. Across these settings,
\methodname{} attains the best mathematical-reasoning averages on both
Qwen3-4B and Qwen3-8B while preserving base-model performance on the out-of-distribution
(OOD) benchmarks.

Our contributions are:
\begin{enumerate}
  \item We introduce \methodname{}, an adaptive iterative repair framework
  that aligns privileged on-policy supervision with the student's evolving
  errors. \methodname{} synthesizes error-specific repair guidance and
  updates it in response to subsequent failed on-policy attempts.
  \item We develop an error-aligned distillation objective that uses newly
  synthesized guidance as privileged teacher context when scoring the
  current failed response. Each supervision region covers the span where the teacher still
  expresses correction, and outcome-aware stage weighting allocates
  supervision across repair stages.
  \item Across Qwen3-4B and Qwen3-8B with self- and external guidance,
  \methodname{} improves average mathematical-reasoning accuracy by up to
  $3.6$ percentage points over the strongest evaluated baseline, while
  keeping average OOD performance near that of the base model.
  Ablations support the roles of error-aligned supervision and stage
  weighting.
\end{enumerate}

\section{Related Work}
\label{sec:related}

\paragraph{Reinforcement learning with verifiable rewards.}
Building on reinforcement learning from human feedback
\citep{schulman2017proximal,ouyang2022training}, reinforcement learning with
verifiable rewards (RLVR) improves language-model reasoning using rewards
from automatically checkable outcomes
\citep{lambert2025tulu,guo2025deepseekr1,yu2025dapo}.
DeepSeekMath introduces Group Relative Policy Optimization (GRPO) for
mathematical reasoning \citep{shao2024deepseekmath}, while DAPO advances
large-scale reasoning training through improved sampling and policy
optimization \citep{yu2025dapo}. Process-level rewards densify this outcome
signal by scoring intermediate steps
\citep{lightman2024lets,wang2024mathshepherd}, yet the supervision remains a
scalar reward rather than a token-level target.
For group-relative methods such as GRPO, difficult samples on which all
rollouts are incorrect provide no relative advantage signal for updating
the policy. To address this limitation, hint-based RL uses auxiliary
guidance to make hard problems more tractable and recover useful training
signals \citep{cao2026hdpo,fan2026lang}.
StepHint guides exploration with partial solution steps
\citep{zhang2025stephint}. NuRL injects abstract hints when independent
attempts fail \citep{chen2026nurl}, while Scaf-GRPO provides progressively
more detailed hints from a pre-generated, fixed hint hierarchy
\citep{zhang2026scafgrpo}.
HiLL further learns hints conditioned on the reasoner's current failures,
seeking to restore informative reward variation while encouraging transfer
to no-hint reasoning \citep{xia2026hill}.
While these approaches use hints to reduce effective task difficulty and
recover outcome-based \emph{reward} signals (from fixed pre-generated
hierarchies or from a learned hinter), \methodname{} synthesizes repair
guidance for each failed response during multi-turn repair and uses it as
privileged teacher context for token-level supervision of the repair.

\paragraph{On-policy (self-) distillation.}
Knowledge distillation trains a student to match a teacher's output
distribution \citep{hinton2015distilling,kim2016sequence}. On-policy
distillation (OPD) instead has the teacher score responses sampled from the
student's own policy
\citep{ross2011reduction,lin2020autoregressive,agarwal2024onpolicy,gu2024minillm,ko2024distillm,lu2026opd},
avoiding the mismatch between fixed demonstrations and the states the
student reaches at deployment. OPSD extends OPD
through in-context self-distillation, where the same model produces teacher
and student distributions under asymmetric contexts
\citep{zhao2026selfdistilled}. In the reference-conditioned setting, the
teacher receives a reference solution that is unavailable to the
student, yielding dense token-level feedback on states that the student
actually visits. Subsequent methods adjust the strength or routing of this
signal using outcomes, reliability, or trajectory structure
\citep{hou2026uniopd,zheng2026scope,zhang2026stepopsd}. Closer to
our setting, several distillation methods anchor supervision at failures
\citep{xu2026relayopd,wang2026motab}.
HINT-SD distills a feedback-conditioned teacher on failure-relevant spans
selected in hindsight \citep{yeo2026hintsd}, and TAPO inserts a
diagnosis-and-correction transition after the learner's first critical error
\citep{huang2026tapo}. Across these methods, the corrected
trajectory or privileged context is constructed in a single pass and does not
respond to how the learner fails again. We focus on how privileged context should be updated from one failed retry
to the next.

\paragraph{Self-correction and iterative refinement.}
Inference-time methods treat an initial response as an intermediate state and
use subsequent feedback to revise it. Self-Refine alternates self-generated
feedback and refinement \citep{madaan2023selfrefine}, while Reflexion converts
task feedback into verbal reflections that guide later trials
\citep{shinn2023reflexion}. Their effectiveness depends on the information in
the feedback, as language models struggle to identify and repair complex
reasoning mistakes without external reference signals \citep{huang2024cannot}.

Training-time methods instead seek to internalize improvement behavior.
Self-training bootstraps from the model's own verified rationales
\citep{zelikman2022star,singh2024beyond,hosseini2024vstar}, RISE fine-tunes on
multi-turn revision trajectories \citep{qu2024recursive}, SCoRe learns
multi-turn self-correction under the model's own distribution via
reinforcement learning \citep{kumar2025training}, and SuperCorrect distills
teacher-generated thought templates and correction traces
\citep{yang2025supercorrect}. \methodname{}
uses iterative repair to construct error-aligned distillation targets on
the student's own failed responses: at each stage, a teacher conditioned
on newly synthesized repair guidance supervises the current response.

\section{Preliminaries: On-Policy Self-Distillation}
\label{sec:method-preliminaries}

Let $q$ denote a problem and $p_\theta$ a trainable student policy that
defines next-token distributions given a conditioning context. On-policy
distillation (OPD) trains the student on states that it actually visits
\citep{agarwal2024onpolicy,lu2026opd}. Given a response $y\sim p_\theta(\cdot\mid q)$
sampled from the current policy, a teacher $p_{\mathrm T}$ scores every
token $y_j$ given its prefix $y_{<j}$ (tokens indexed from $0$), and OPD
minimizes the token-level divergence
\begin{equation}
  \mathcal{L}_{\mathrm{OPD}}(q,y)
  =
  \frac{1}{|y|}
  \sum_{j=0}^{|y|-1}
  D\!\left(
    p_{\mathrm T}(\cdot\mid q,y_{<j})
    \,\big\|\,
    p_\theta(\cdot\mid q,y_{<j})
  \right).
  \label{eq:opd}
\end{equation}
Because the response is sampled from $p_\theta$, the teacher provides dense
feedback on prefixes generated by the evolving student. In OPD, the teacher
is typically a separate, more capable model.

On-policy self-distillation (OPSD) further initializes the student and the
teacher from the same model, removing the need for a separately trained
teacher \citep{zhao2026selfdistilled}. Providing the reference solution $z$
as additional input to the teacher gives it information that the student
cannot access. Here the teacher $p_{\mathrm T}$ is instantiated as the
frozen initial backbone $\bar{p}$. For an on-policy response
$y\sim p_\theta(\cdot\mid q)$, distillation compares the two branches at
the same student-generated prefixes $y_{<j}$. The teacher conditions on
$(q,z,y_{<j})$, the student conditions on $(q,y_{<j})$, and the OPSD loss
is
\begin{equation}
  \mathcal{L}_{\mathrm{OPSD}}(q,z,y)
  =
  \frac{1}{|y|}
  \sum_{j=0}^{|y|-1}
  D\!\left(
    p_{\mathrm T}(\cdot\mid q,z,y_{<j})
    \,\big\|\,
    p_\theta(\cdot\mid q,y_{<j})
  \right),
  \label{eq:opsd-loss}
\end{equation}
where only the teacher branch observes $z$.
Following OPSD, we instantiate $D$ as the forward KL divergence
$\KL\!\left(p_{\mathrm T}\,\Vert\,p_\theta\right)$ and apply pointwise
clipping, which caps each vocabulary entry's divergence contribution so that
a few high-divergence tokens cannot dominate the update.

In OPSD, in-context learning thus converts the reference solution into
fine-grained supervision over the probability of each token in the
student's on-policy response.
However, under this construction, the teacher must extract relevant information
from the complete reference solution and relate it to the student's current
reasoning response. This process is performed implicitly and does not
explicitly teach the student how to transition from its current error state to
a valid solution state.

\section{\methodname: Adaptive Iterative Repair for On-Policy Distillation}
\label{sec:method}

\begin{figure}[t]
\centering
\includegraphics[width=\textwidth]{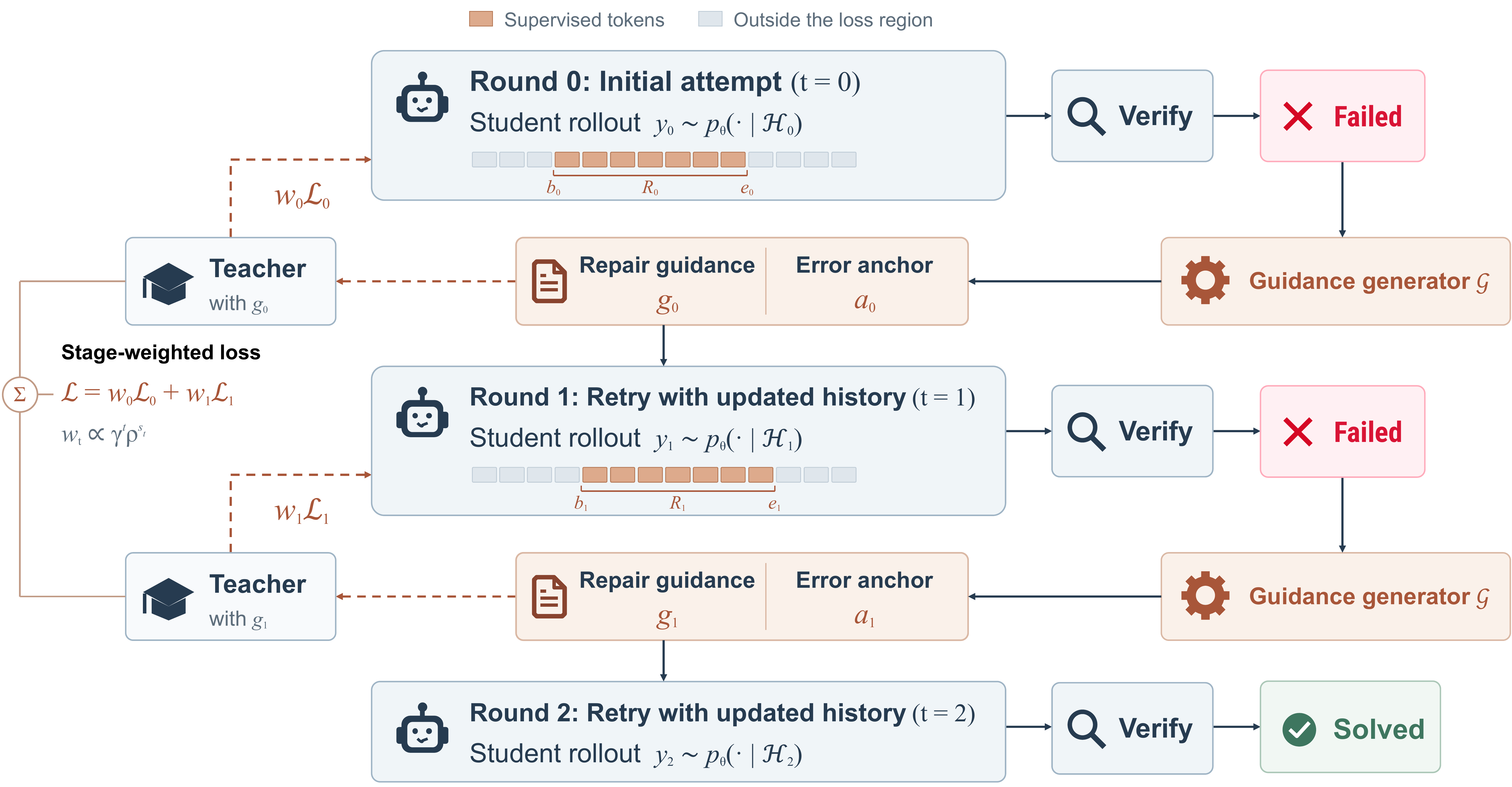}
\caption{\textbf{Overview of \methodname.} When an on-policy rollout fails
verification, the guidance generator $\mathcal{G}$ quotes the first flawed
step as an error anchor $a_t$ and synthesizes repair guidance $g_t$. A
frozen teacher conditioned on $g_t$ then supervises the student over the
error-aligned repair region $R_t$, and the student retries with the
guidance in context. A verified retry ends the trajectory, and outcome-aware stage
weighting combines the stage losses with weights
$w_t\propto\gamma^{t}\rho^{s_t}$.}
\label{fig:hero}
\end{figure}

\methodname{} makes explicit the extraction step that OPSD leaves to the
teacher (Figure~\ref{fig:hero}). A guidance generator reads the reference
solution and writes repair guidance for the student's current error, and
the teacher receives this guidance in place of the full reference
solution. The rest of the method answers three questions: how the guidance
is updated across repair rounds (Section~\ref{sec:method-multiround}),
which tokens of each failed response the teacher supervises
(Section~\ref{sec:method-scaffold}), and how the resulting stage losses
are combined (Section~\ref{sec:method-objective}).

\subsection{Multi-round adaptive repair}
\label{sec:method-multiround}

\methodname{} performs up to $K$ repair rounds, adapting the guidance to
the student's latest failed response. Let $\mathcal{H}_t$ denote the
student-visible dialogue history at stage $t$, with $\mathcal{H}_0$
containing only the problem $q$. The student generates an on-policy
response
\begin{equation}
  y_t \sim p_\theta(\cdot\mid\mathcal{H}_t).
  \label{eq:stage-rollout}
\end{equation}
If $y_t$ fails verification by an outcome verifier $V$, the guidance
generator $\mathcal{G}$ uses
the reference solution $z$ to diagnose the current failure and produce
\begin{equation}
  (a_t,g_t)
  =
  \mathcal{G}(q,z,\mathcal{H}_t,y_t).
  \label{eq:scaffold-generation}
\end{equation}
Here, $a_t$ quotes the identified first flawed step verbatim, and $g_t$
provides concise guidance for correcting it. Rule-based checks filter out
guidance that reveals the final answer
(Appendix~\ref{sec:appendix-prompts} details the rules).

To advance the repair process, we prompt the student to repair the previous
erroneous response $y_t$ according to $g_t$:
\begin{equation}
  \mathcal{H}_{t+1}
  =
  \operatorname{Update}(\mathcal{H}_t,y_t,g_t).
  \label{eq:history-update}
\end{equation}
$\operatorname{Update}$ appends $y_t$ to the history as the model's own
output turn, and appends $g_t$, rendered with a repair instruction, as the
next input turn. If budget remains, the student then generates the retry
$y_{t+1}\sim p_\theta(\cdot\mid\mathcal{H}_{t+1})$; if the retry remains
incorrect, $\mathcal{G}$ re-localizes the error in $y_{t+1}$ and generates
updated guidance. The process ends when a response passes verification or
the budget is exhausted. Throughout
this process, the privileged context keeps adapting to the student's
current error state. Appendix~\ref{sec:appendix-case} presents worked
repair examples from training.

\subsection{Error-aligned privileged distillation}
\label{sec:method-scaffold}

At stage $t$, the teacher and student score the same failed response $y_t$
under the shared dialogue history $\mathcal{H}_t$. The frozen teacher
$p_{\mathrm T}$ additionally receives the newly generated guidance $g_t$, which
is withheld from the student when scoring $y_t$.

To cover both the current error and the potentially recoverable region that
follows, the supervision region starts at the beginning of the error
region. We exactly match $a_t$ against $y_t$ to obtain the start position
$b_t$. Supervising the error region alone would lose this recoverable
region, so we rely on the corrective signal from the teacher to determine
the endpoint of the supervision region. At position $j$, we measure this
signal as the total positive teacher--student probability difference over
alternatives to the observed token:
\begin{equation}
  C_{t,j}
  =
  \sum_{v\neq y_{t,j}}
  \left[
    p_{\mathrm T}\!\left(v\mid \mathcal{H}_t,g_t,y_{t,<j}\right)
    -
    p_\theta\!\left(v\mid \mathcal{H}_t,y_{t,<j}\right)
  \right]_{+}.
  \label{eq:correction-mass}
\end{equation}
Both distributions are already computed for the stage loss, so measuring
this signal introduces no additional computational overhead.

To avoid ending the region within a reasoning block, we partition the
suffix of $y_t$ starting at $b_t$ at blank-line boundaries. Let
$\mathcal{E}_t$ denote the resulting block endpoints, including the response
end $|y_t|$. For a
tolerance $\epsilon\in(0,1)$, we select the earliest endpoint retaining at
least a $1-\epsilon$ fraction of the suffix's total corrective signal:
\begin{equation}
  e_t
  =
  \min\Bigl\{
    r\in\mathcal{E}_t:
    \textstyle\sum_{j=b_t}^{r-1} C_{t,j}
    \geq
    (1-\epsilon)\textstyle\sum_{j=b_t}^{|y_t|-1} C_{t,j}
  \Bigr\}.
  \label{eq:endpoint}
\end{equation}
By construction, the discarded tail carries at most an $\epsilon$ fraction
of the suffix's total corrective signal. If the total signal is zero, we
retain the full suffix. Following
Eq.~\ref{eq:opsd-loss}, the stage loss averages the divergence over the
selected repair region $R_t=y_t[b_t:e_t]$, i.e., tokens $b_t$ to $e_t-1$:
\begin{equation}
  \mathcal{L}_{t}
  =
  \frac{1}{e_t-b_t}
  \sum_{j=b_t}^{e_t-1}
  D\!\left(
    p_{\mathrm T}\!\left(\cdot\mid \mathcal{H}_t,g_t,y_{t,<j}\right)
    \,\big\|\,
    p_\theta\!\left(\cdot\mid \mathcal{H}_t,y_{t,<j}\right)
  \right).
  \label{eq:stage-loss}
\end{equation}
Gradients flow only through the student branch.

\subsection{Outcome-aware stage weighting}
\label{sec:method-objective}

Training samples contain different numbers of repair stages, and directly
summing stage losses would grant samples with more repair rounds a larger
total weight. To avoid this, we introduce
\emph{outcome-aware stage weighting}. The scheme allocates the stage
weights within each supervised trajectory by stage depth
and the outcome of the retry immediately following each stage.

This allocation rests on two assumptions about the repair process. First,
imperfect retries accumulate noise across stages. The later a stage is, the
more failed responses and repair guidance accumulate in its context and the
less reliable its supervision signal becomes, so the weights should lean
toward earlier stages. Second, whether the immediate retry passes
verification reflects the stage's supervision quality. A successful retry
indicates that the stage's repair
guidance pointed to a valid repair, and such a stage should receive an
additional weight bonus.

Concretely, for trajectory $n$ with
final supervised stage $T_n$ and stage losses $\mathcal{L}_{n,t}$
(Eq.~\ref{eq:stage-loss}), let $s_{n,t}=\mathds{1}[V(y_{n,t+1})=1]$
indicate whether the on-policy retry immediately following stage $t$ passes
the outcome verifier $V$, with $s_{n,t}=0$ when no retry is available
(after the final round). The stage weights are
\begin{equation}
\begin{gathered}
  w_{n,t}
  =
  \frac{\widetilde{w}_{n,t}}
       {\sum_{k=0}^{T_n}\widetilde{w}_{n,k}},
  \qquad
  \widetilde{w}_{n,t}
  =
  \gamma^{t}\rho^{s_{n,t}},
  \\
  \gamma\in(0,1],\qquad \rho\geq 1.
\end{gathered}
  \label{eq:normalized-stage-weight}
\end{equation}
The temporal factor $\gamma^{t}$ (with $\gamma<1$) decays the weights
with stage depth, the outcome bonus $\rho^{s_{n,t}}$ rewards a verified
immediate retry, and the normalization gives each supervised trajectory
unit total weight.

For a minibatch of repair trajectories $\mathcal{B}$, the final
objective is
\begin{equation}
  \mathcal{L}_{\methodname}
  =
  \frac{1}{|\mathcal{B}|}
  \sum_{n\in\mathcal{B}}
  \sum_{t=0}^{T_n}
  w_{n,t}\mathcal{L}_{n,t}.
  \label{eq:air-opd-objective}
\end{equation}
The complete algorithm is summarized in
Appendix~\ref{sec:appendix-algorithm}.

\section{Experimental Setup}
\label{sec:setup}
\label{sec:experimental-setup}

\subsection{Models, training data, and baselines}

We use two Qwen variants of different scales, Qwen3-4B and Qwen3-8B
\citep{yang2025qwen3}. We train on the DAPO-Math-17K dataset
\citep{yu2025dapo}, generating the reference solution for each problem with
GPT-5.5 and retaining the 15{,}650 problems whose generated
solution verifies against the dataset answer. Outcome verification at
every repair round uses rule-based answer matching. We compare
\emph{self-guidance} generated by the current student policy with \emph{external
guidance} generated by a larger model (Qwen3.5-27B). The main baselines are
the corresponding base model and three trained methods. Supervised
fine-tuning (SFT) trains on the reference solutions with a cross-entropy
loss. GRPO optimizes rule-based outcome rewards over groups of on-policy
rollouts. OPSD distills a reference-conditioned teacher on the student's
own rollouts. All trained baselines are our
reproductions on the same training corpus and backbones
(Appendix~\ref{sec:appendix-evaluation}).

\subsection{Evaluation benchmarks and metrics}

We evaluate models along two dimensions, mathematical reasoning and
out-of-distribution generalization. Mathematical reasoning is evaluated on
AIME24, AIME25, and HMMT25 \citep{hmmt2025}. Each mathematical
benchmark contains 30 problems. We report
per-benchmark accuracy as Avg@12, with \emph{Math Avg.} the unweighted mean of
the three benchmark scores. Out-of-distribution evaluation uses MMLU-Pro
\citep{wang2024mmlupro} and GPQA \citep{rein2024gpqa} to assess transfer beyond the mathematical training distribution,
reporting accuracy as Avg@4, with \emph{OOD Avg.} the mean of the two
benchmarks. During evaluation, no repair guidance is added to the model input.

\subsection{Implementation details}
\label{sec:setup-implementation}

We adopt the OPSD training recipe \citep{zhao2026selfdistilled}, using the
frozen initial backbone with its adapters disabled as the distillation
teacher. During training, the student operates in non-thinking mode and the
teacher in thinking mode. Optimization uses rank-64 LoRA adapters
\citep{hu2022lora} and AdamW \citep{loshchilov2019decoupled} with a learning
rate of $5\times10^{-6}$ and an effective batch size of 32.

\methodname{} is built on TRL \citep{vonwerra2020trl} and trains for 100
optimizer steps. Each
problem allows at most $K=3$ adaptive repair rounds, and each round samples a
single on-policy rollout from the current student (temperature 1.1,
top-$p$ 0.95), capped at 8{,}192 tokens. Sampler weights are synchronized
after every optimizer step. The default stage weighting is
$(\gamma,\rho)=(0.8,1.5)$, and the endpoint budget is $\epsilon=0.1$.
Both guidance sources are generated in thinking mode (temperature 0.7,
top-$p$ 0.9) and filtered by the same validation rules.
All experiments are conducted on a single node with eight H100 GPUs.
Appendix~\ref{sec:appendix-evaluation} documents the complete training and
evaluation configuration.

\section{Main Results}
\label{sec:main-results}

\textbf{\methodname{} achieves the best mathematical reasoning at both
scales.} Table~\ref{tab:main-results} compares \methodname{} against the
base model, SFT, GRPO, and OPSD. Both guidance settings surpass every
baseline on Math Avg., and external guidance, where a stronger generator
supplies higher-quality repair guidance, exceeds the strongest baseline,
OPSD, by 3.6 and 2.7 points at 4B and 8B, respectively. The
improvement is consistent across benchmarks. \methodname{} variants
attain the best score on every mathematical benchmark at both scales, with
pronounced gains on HMMT25.

\begin{table*}[t]
\centering
\caption{\textbf{Main results on mathematical-reasoning (Avg@12) and out-of-distribution (Avg@4) benchmarks.} Both \methodname{} variants surpass all baselines on Math Avg. at each scale while preserving base-model performance on the OOD group. \textsc{Self-G}/\textsc{External-G} generate repair guidance with the current student policy and Qwen3.5-27B, respectively. \textbf{Bold}/\underline{underlined}: best/second-best within each block.}
\label{tab:main-results}
\small
\setlength{\tabcolsep}{2.8pt}
\renewcommand{\arraystretch}{1.05}
\begin{tabular}{lccccccc}
\toprule
& \multicolumn{4}{c}{Mathematical Reasoning} &
  \multicolumn{3}{c}{Out-of-Distribution} \\
\cmidrule(lr){2-5}\cmidrule(lr){6-8}
Method & AIME24 & AIME25 & HMMT25 & Math Avg. & MMLU-Pro & GPQA & OOD Avg. \\
\midrule
\multicolumn{8}{l}{\textit{Qwen3-4B}} \\
Base model & 75.0 & 66.4 & 42.2 & 61.2 & 51.5 & \textbf{62.9} & \textbf{57.2} \\
SFT & 70.6 & 61.4 & 42.8 & 58.2 & 43.1 & 52.0 & 47.6 \\
GRPO & 75.6 & 67.5 & 43.9 & 62.3 & 51.3 & \underline{62.4} & 56.8 \\
OPSD & 76.4 & 69.4 & 43.3 & 63.1 & 50.4 & 60.2 & 55.3 \\
\methodname{} (\textsc{Self-G}) & \underline{76.7} & \underline{73.1} & \textbf{47.8} & \underline{65.8} & \textbf{52.5} & 61.4 & \underline{56.9} \\
\methodname{} (\textsc{External-G}) & \textbf{80.0} & \textbf{73.3} & \underline{46.7} & \textbf{66.7} & \underline{51.7} & 61.9 & 56.8 \\
\midrule
\multicolumn{8}{l}{\textit{Qwen3-8B}} \\
Base model & 75.8 & 65.6 & 43.9 & 61.8 & 57.9 & \underline{64.1} & \underline{61.0} \\
SFT & 75.3 & 63.6 & 43.6 & 60.8 & 46.5 & 55.8 & 51.2 \\
GRPO & 76.9 & 69.2 & 45.8 & 64.0 & 56.6 & \textbf{64.5} & 60.6 \\
OPSD & 77.8 & 69.7 & 46.7 & 64.7 & 57.4 & 61.7 & 59.6 \\
\methodname{} (\textsc{Self-G}) & \underline{78.3} & \underline{72.5} & \textbf{49.7} & \underline{66.9} & \textbf{58.3} & 63.5 & 60.9 \\
\methodname{} (\textsc{External-G}) & \textbf{79.7} & \textbf{73.1} & \underline{49.4} & \textbf{67.4} & \underline{58.0} & \underline{64.1} & \textbf{61.1} \\
\bottomrule
\end{tabular}
\end{table*}

\textbf{Self-guidance alone accounts for most of the improvement.} Without
introducing any external model, the self-guided variant already improves
over OPSD by 2.8 and 2.1 Math Avg. points. External guidance further
improves the Math Avg. The additional gain from external guidance is
smaller at 8B than at 4B, possibly because the larger student generates
better repair guidance itself.

\textbf{\methodname{} attains stronger mathematical reasoning while
preserving near-base OOD performance.} SFT lowers the OOD average by 9.6
and 9.8 points at 4B and 8B, consistent with forgetting under task-specific
fine-tuning \citep{kirkpatrick2017overcoming,kotha2024understanding}, and
OPSD also regresses at both scales. In contrast, \methodname{} stays within
0.4 points of the base model at 4B and within 0.1 points at 8B. It is the only
evaluated reference-based training method that maintains a near-base OOD
average at both scales. GRPO preserves OOD performance comparably, but its
Math Avg. trails \methodname{} by $2.9$--$4.4$ points. One
possible explanation lies in the granularity of supervision. RLVR provides
only scalar feedback, and full-response distillation supervises every
token, whereas \methodname{} provides corrective signals only on regions
aligned with the student's errors. This selective supervision may limit
over-specialization to the mathematical training distribution while
preserving the gains of distillation.

\section{Ablations and Analysis}
\label{sec:analysis}

All ablations start from the default Qwen3-4B configuration of
Section~\ref{sec:setup-implementation} and report the in-domain Math Avg.
We organize the analysis into three
parts. Section~\ref{sec:analysis-components} controls for
failure-only sample selection and ablates the supervision region and the
stage weighting of the objective,
Section~\ref{sec:analysis-rounds} measures how large a repair
budget is needed, and Section~\ref{sec:analysis-dynamics} traces how
capability evolves over the course of training relative to GRPO and OPSD.

\subsection{Effect of sample selection, supervision region, and stage weighting}
\label{sec:analysis-components}

In Table~\ref{tab:ablation-components}, the two OPSD rows vary which
rollouts are supervised, and each remaining row changes one component of
\methodname{}, the supervision region (Section~\ref{sec:method-scaffold})
or the stage weighting (Section~\ref{sec:method-objective}), while keeping
the rest of the default configuration.

\begin{wraptable}{r}{0.48\textwidth}
\vspace{-12pt}
\centering
\captionsetup{font=small}
\setlength{\tabcolsep}{4pt}
\renewcommand{\arraystretch}{1.05}
\caption{\textbf{Ablation results (Qwen3-4B, \textsc{Self-G}).} We test failure-only
sample selection for OPSD and ablate the supervision region and the
stage-weighting components of \methodname{}.}
\label{tab:ablation-components}
\small
\begin{tabular}{lc}
\toprule
Setting & Math Avg. \\
\midrule
\methodname{} & 65.8 \\
\midrule
\multicolumn{2}{l}{\textit{Sample selection (OPSD)}} \\
\quad All rollouts & 63.1 \\
\quad Failed rollouts only & 62.5 \\
\midrule
\multicolumn{2}{l}{\textit{Supervision region}} \\
\quad w/o error alignment & 64.7 \\
\quad Error block only & 64.2 \\
\midrule
\multicolumn{2}{l}{\textit{Stage weighting $(\gamma,\rho)$}} \\
\quad Uniform $(1,\,1)$ & 65.0 \\
\quad Temporal factor only $(0.8,\,1)$ & 65.4 \\
\quad Outcome bonus only $(1,\,1.5)$ & 65.4 \\
\bottomrule
\end{tabular}
\vspace{-8pt}
\end{wraptable}

\textbf{Sample selection.} \methodname{} supervises only failed rollouts,
so its gains could stem from this selection alone. Applying the OPSD loss
only to failed rollouts yields 62.5, 0.6 points below standard OPSD and
3.3 below \methodname{}. The gains thus depend on how \methodname{}
supervises each failure, not on which rollouts it selects.

\textbf{Supervision region.} We compare
the default repair region $R_t$ with two ablations. \emph{W/o error
alignment} supervises every token of the failed response in place of this
region, and \emph{error block only} ends the region at the boundary of the
first reasoning block. Both ablations lower the Math Avg. Widening
supervision to the full response costs 1.1 points, and narrowing it to the
error block alone costs 1.6. We suspect that full-response supervision
dilutes the signal with tokens the repair guidance does not target, while
the error block alone omits downstream reasoning that may also require
correction. These results support supervising both the identified error
region and the selected downstream reasoning.

\begin{wrapfigure}{r}{0.46\textwidth}
\vspace{-12pt}
\centering
\includegraphics[width=\linewidth]{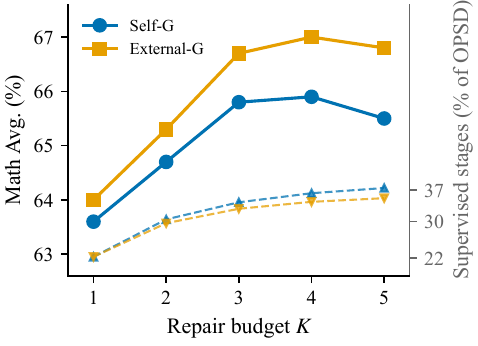}
\caption{\textbf{Effect of the repair budget $K$ (Qwen3-4B).} Math Avg.
(left axis) rises through the default $K=3$ and then flattens, while each
variant's number of supervised stages (dashed, right axis; \% of the
samples OPSD supervises) grows with diminishing increments.}
\label{fig:budget-sweep}
\vspace{-8pt}
\end{wrapfigure}
\textbf{Stage weighting.} We ablate the temporal factor $\gamma^{t}$ and
the outcome bonus $\rho^{s_{n,t}}$ of
Eq.~\ref{eq:normalized-stage-weight}.
\emph{Uniform} $(1,1)$ removes both factors and weights all stages equally,
while \emph{temporal factor only} $(0.8,1)$ and \emph{outcome
bonus only} $(1,1.5)$ each restore a single factor. Either factor alone
already improves over uniform weighting, and combining the two performs
best. This indicates that the weights should combine a preference for
earlier stages with feedback from the immediate retry outcome rather than
allocate supervision by a single criterion.

\subsection{Effect of the repair budget}
\label{sec:analysis-rounds}

We vary the maximum number of repair rounds $K\in\{1,\ldots,5\}$ for both
guidance settings. Figure~\ref{fig:budget-sweep} traces Math Avg. as the
budget grows alongside the number of supervised stages. Both variants
already surpass OPSD with a single repair round, and the overall trend
rises and then flattens. At $K=1$ every supervised trajectory contains
exactly one stage, so this gain does not depend on additional repair
rounds. $K=1$ also exceeds failure-only OPSD by 1.1--1.5 points
(Table~\ref{tab:ablation-components}), suggesting that error-aligned repair
supervision provides a stronger learning signal than
reference-conditioned full-response distillation. Raising the budget to $K=3$ brings further
gains, which confirms the value of continuing to supply fresh repair
signals for unresolved failures during training. Beyond this point both
curves show diminishing returns. $K=4$ adds only a small improvement, and
$K=5$ falls slightly below $K=4$. We suspect that stages added at deep
rounds carry accumulated noise, as their contexts pile up failed attempts
and guidance, and their less reliable supervision begins to offset the
gains.

The dashed curves show that the number of supervised stages keeps
increasing with $K$ but with shrinking increments, as later rounds activate
only for trajectories that remain incorrect and yield validated repair
guidance rather than for all trajectories. The count rises from
22.2\% of the samples OPSD supervises at $K=1$ to 35.2--37.4\% at
$K=5$. We keep $K=3$ as the practical default. It captures most of the
observed gains while requiring fewer supervised stages than larger budgets.
Comparing the two dashed curves further shows that external guidance adds
fewer supervised stages per extra round than self-guidance, possibly
because higher-quality external guidance resolves more failures in earlier
rounds.

\begin{figure}[t]
\centering
\includegraphics[width=\textwidth]{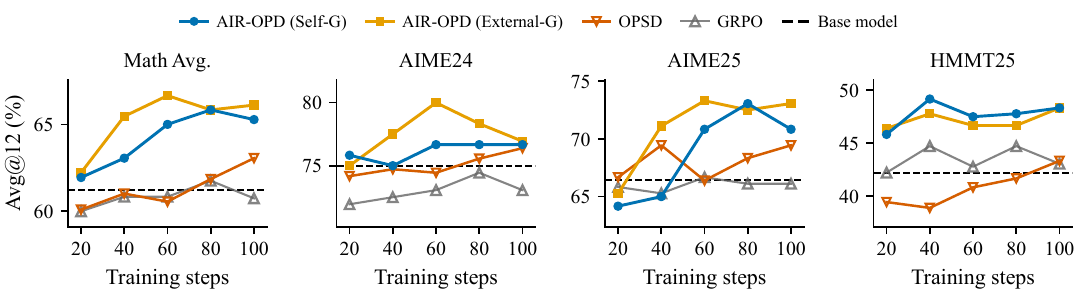}
\caption{\textbf{Training dynamics on the mathematical benchmarks
(Qwen3-4B).} Both \methodname{} variants lead OPSD and GRPO on the
three-benchmark average from the first checkpoint onward. Curves show
Avg@12 at checkpoints saved every 20 optimizer steps, and dashed lines mark
the base model.}
\label{fig:training-dynamics}
\end{figure}

\subsection{Training dynamics}
\label{sec:analysis-dynamics}

Figure~\ref{fig:training-dynamics} plots the three-benchmark average and
per-benchmark accuracy over the checkpoints of both \methodname{} variants,
OPSD, and GRPO, saved every 20 optimizer steps, with the base model as a
fixed reference. On the three-benchmark
average, both \methodname{} variants lead both baselines at every checkpoint,
and external guidance matches or exceeds the self-guided average throughout.
\methodname{} also converges substantially faster. By step 40, both variants
already reach the score OPSD attains at step 100.
This faster convergence also uses much less supervision. At
the default $K=3$, the number of stages supervised by \methodname{} is
approximately one third of the number of samples supervised by OPSD
(Section~\ref{sec:analysis-rounds}), with each stage covering an
error-aligned repair region rather than a full response. This suggests
that supervision concentrated on erroneous regions yields more learning
signal per supervised token.

\section{Conclusion}
\label{sec:conclusion}

We introduced \methodname{}, an adaptive iterative repair framework for
on-policy distillation that addresses the solution-conditioned shortcut
risk of reference-privileged supervision. Instead of conditioning the teacher on a
static reference solution, a guidance generator synthesizes answer-free
repair guidance that evolves with the student's errors, the teacher distills
each repair over an error-aligned region of the latest failed response, and
outcome-aware stage weighting allocates supervision across rounds. Across two model scales, both guidance sources surpass strong RL and
distillation baselines on mathematical reasoning while preserving
base-model performance out of distribution, and self-guidance alone accounts for
most of the improvement. Our analysis further shows that the gains emerge from the
first checkpoints of training, and that a small repair budget captures
most of the gains observed in our sweep while supervising roughly a third
as many samples as OPSD.

\bibliography{references}

\begin{thebibliography}{53}
\providecommand{\natexlab}[1]{#1}
\providecommand{\url}[1]{\texttt{#1}}
\expandafter\ifx\csname urlstyle\endcsname\relax
  \providecommand{\doi}[1]{doi: #1}\else
  \providecommand{\doi}{doi: \begingroup \urlstyle{rm}\Url}\fi

\bibitem[Agarwal et~al.(2024)Agarwal, Vieillard, Zhou, Stanczyk, Ramos, Geist,
  and Bachem]{agarwal2024onpolicy}
Rishabh Agarwal, Nino Vieillard, Yongchao Zhou, Piotr Stanczyk, Sabela Ramos,
  Matthieu Geist, and Olivier Bachem.
\newblock On-policy distillation of language models: Learning from
  self-generated mistakes.
\newblock In \emph{International Conference on Learning Representations
  (ICLR)}, 2024.
\newblock URL \url{https://openreview.net/forum?id=3zKtaqxLhW}.

\bibitem[Cao et~al.(2026)Cao, Wu, Zhong, Li, Li, Ye, and Zhu]{cao2026hdpo}
Zhiyu Cao, Kaixin Wu, Mingjie Zhong, Peifeng Li, Xiaobo Li, Can Ye, and
  Qiaoming Zhu.
\newblock Hint-guided diversified policy optimization for {LLM} reasoning.
\newblock \emph{arXiv preprint arXiv:2606.03021}, 2026.
\newblock URL \url{https://arxiv.org/abs/2606.03021}.

\bibitem[Chen et~al.(2026)Chen, Peng, Choubey, Huang, Zhang, Bansal, and
  Wu]{chen2026nurl}
Justin Chih-Yao Chen, Becky~Xiangyu Peng, Prafulla~Kumar Choubey, Kung-Hsiang
  Huang, Jiaxin Zhang, Mohit Bansal, and Chien-Sheng Wu.
\newblock Nudging the boundaries of {LLM} reasoning.
\newblock In \emph{International Conference on Learning Representations
  (ICLR)}, 2026.
\newblock URL \url{https://openreview.net/forum?id=hfNnQHkTtv}.

\bibitem[Dao(2024)]{dao2024flashattention}
Tri Dao.
\newblock {FlashAttention-2}: Faster attention with better parallelism and work
  partitioning.
\newblock In \emph{International Conference on Learning Representations
  (ICLR)}, 2024.
\newblock URL \url{https://openreview.net/forum?id=mZn2Xyh9Ec}.

\bibitem[Fan et~al.(2026)Fan, Li, Li, Wang, Mu, Yang, Chen, Weng, Wang, Cai,
  Zhu, and Xiao]{fan2026lang}
Yuchun Fan, Bei Li, Peiguang Li, Yilin Wang, Yongyu Mu, Jian Yang, Xin Chen,
  Rongxiang Weng, Jingang Wang, Xunliang Cai, Jingbo Zhu, and Tong Xiao.
\newblock {LANG}: Reinforcement learning for multilingual reasoning with
  language-adaptive hint guidance.
\newblock In \emph{Proceedings of the 64th Annual Meeting of the Association
  for Computational Linguistics (Volume 1: Long Papers)}, pp.\  43838--43866,
  2026.
\newblock URL \url{https://aclanthology.org/2026.acl-long.2029/}.

\bibitem[Gu et~al.(2024)Gu, Dong, Wei, and Huang]{gu2024minillm}
Yuxian Gu, Li~Dong, Furu Wei, and Minlie Huang.
\newblock {MiniLLM}: Knowledge distillation of large language models.
\newblock In \emph{International Conference on Learning Representations
  (ICLR)}, 2024.
\newblock URL \url{https://openreview.net/forum?id=5h0qf7IBZZ}.

\bibitem[Guo et~al.(2025)Guo, Yang, Zhang, Song, et~al.]{guo2025deepseekr1}
Daya Guo, Dejian Yang, Haowei Zhang, Junxiao Song, et~al.
\newblock {DeepSeek-R1} incentivizes reasoning in {LLMs} through reinforcement
  learning.
\newblock \emph{Nature}, 645\penalty0 (8081):\penalty0 633--638, 2025.
\newblock \doi{10.1038/s41586-025-09422-z}.
\newblock URL \url{https://doi.org/10.1038/s41586-025-09422-z}.

\bibitem[Hinton et~al.(2015)Hinton, Vinyals, and Dean]{hinton2015distilling}
Geoffrey Hinton, Oriol Vinyals, and Jeff Dean.
\newblock Distilling the knowledge in a neural network.
\newblock \emph{arXiv preprint arXiv:1503.02531}, 2015.
\newblock URL \url{https://arxiv.org/abs/1503.02531}.

\bibitem[{HMMT}(2025)]{hmmt2025}
{HMMT}.
\newblock Harvard--{MIT} mathematics tournament ({HMMT}) february 2025.
\newblock \url{https://www.hmmt.org/}, 2025.
\newblock Accessed 2026-09.

\bibitem[Hosseini et~al.(2024)Hosseini, Yuan, Malkin, Courville, Sordoni, and
  Agarwal]{hosseini2024vstar}
Arian Hosseini, Xingdi Yuan, Nikolay Malkin, Aaron Courville, Alessandro
  Sordoni, and Rishabh Agarwal.
\newblock {V-STaR}: Training verifiers for self-taught reasoners.
\newblock In \emph{Conference on Language Modeling (COLM)}, 2024.
\newblock URL \url{https://openreview.net/forum?id=stmqBSW2dV}.

\bibitem[Hou et~al.(2026)Hou, Peng, Wang, Ruan, Zhang, Zhou, Gao, Chen, Wang,
  Yang, Zhang, Tian, Hu, Yang, Wu, and Fan]{hou2026uniopd}
Wenjin Hou, Shangpin Peng, Weinong Wang, Zheng Ruan, Yue Zhang, Zhenglin Zhou,
  Mingqi Gao, Yifei Chen, Kaiqi Wang, Hongming Yang, Chengquan Zhang, Zhuotao
  Tian, Han Hu, Yi~Yang, Fei Wu, and Hehe Fan.
\newblock Uni-opd: Unifying on-policy distillation with a dual-perspective
  recipe.
\newblock \emph{arXiv preprint arXiv:2605.03677}, 2026.
\newblock URL \url{https://arxiv.org/abs/2605.03677}.

\bibitem[Hu et~al.(2022)Hu, Shen, Wallis, Allen-Zhu, Li, Wang, Wang, and
  Chen]{hu2022lora}
Edward~J. Hu, Yelong Shen, Phillip Wallis, Zeyuan Allen-Zhu, Yuanzhi Li, Shean
  Wang, Lu~Wang, and Weizhu Chen.
\newblock {LoRA}: Low-rank adaptation of large language models.
\newblock In \emph{International Conference on Learning Representations
  (ICLR)}, 2022.
\newblock URL \url{https://openreview.net/forum?id=nZeVKeeFYf9}.

\bibitem[Huang et~al.(2024)Huang, Chen, Mishra, Zheng, Yu, Song, and
  Zhou]{huang2024cannot}
Jie Huang, Xinyun Chen, Swaroop Mishra, Huaixiu~Steven Zheng, Adams~Wei Yu,
  Xinying Song, and Denny Zhou.
\newblock Large language models cannot self-correct reasoning yet.
\newblock In \emph{International Conference on Learning Representations
  (ICLR)}, 2024.
\newblock URL \url{https://openreview.net/forum?id=IkmD3fKBPQ}.

\bibitem[Huang et~al.(2026)Huang, Gao, Dong, Chen, Luo, Qin, Wang, Yang, and
  Jiang]{huang2026tapo}
Zhilin Huang, Hang Gao, Ziqiang Dong, Yuan Chen, Yifeng Luo, Chujun Qin, Jingyi
  Wang, Yang Yang, and Guanjun Jiang.
\newblock Learning from your own mistakes: Constructing learnable
  micro-reflective trajectories for self-distillation.
\newblock \emph{arXiv preprint arXiv:2606.18844}, 2026.
\newblock URL \url{https://arxiv.org/abs/2606.18844}.

\bibitem[Kim \& Rush(2016)Kim and Rush]{kim2016sequence}
Yoon Kim and Alexander~M. Rush.
\newblock Sequence-level knowledge distillation.
\newblock In \emph{Proceedings of the 2016 Conference on Empirical Methods in
  Natural Language Processing (EMNLP)}, pp.\  1317--1327, 2016.
\newblock URL \url{https://aclanthology.org/D16-1139/}.

\bibitem[Kirkpatrick et~al.(2017)Kirkpatrick, Pascanu, Rabinowitz, Veness,
  Desjardins, Rusu, Milan, Quan, Ramalho, Grabska-Barwinska,
  et~al.]{kirkpatrick2017overcoming}
James Kirkpatrick, Razvan Pascanu, Neil Rabinowitz, Joel Veness, Guillaume
  Desjardins, Andrei~A. Rusu, Kieran Milan, John Quan, Tiago Ramalho, Agnieszka
  Grabska-Barwinska, et~al.
\newblock Overcoming catastrophic forgetting in neural networks.
\newblock \emph{Proceedings of the National Academy of Sciences (PNAS)},
  114\penalty0 (13):\penalty0 3521--3526, 2017.
\newblock \doi{10.1073/pnas.1611835114}.

\bibitem[Ko et~al.(2024)Ko, Kim, Chen, and Yun]{ko2024distillm}
Jongwoo Ko, Sungnyun Kim, Tianyi Chen, and Se-Young Yun.
\newblock {DistiLLM}: Towards streamlined distillation for large language
  models.
\newblock In \emph{Proceedings of the 41st International Conference on Machine
  Learning (ICML)}, 2024.
\newblock URL \url{https://proceedings.mlr.press/v235/ko24c.html}.

\bibitem[Kotha et~al.(2024)Kotha, Springer, and
  Raghunathan]{kotha2024understanding}
Suhas Kotha, Jacob~Mitchell Springer, and Aditi Raghunathan.
\newblock Understanding catastrophic forgetting in language models via implicit
  inference.
\newblock In \emph{International Conference on Learning Representations
  (ICLR)}, 2024.
\newblock URL \url{https://openreview.net/forum?id=VrHiF2hsrm}.

\bibitem[Kumar et~al.(2025)Kumar, Zhuang, Agarwal, Su, Co-Reyes, Singh, Baumli,
  Iqbal, Bishop, Roelofs, Zhang, McKinney, Shrivastava, Paduraru, Tucker,
  Precup, Behbahani, and Faust]{kumar2025training}
Aviral Kumar, Vincent Zhuang, Rishabh Agarwal, Yi~Su, John~D. Co-Reyes, Avi
  Singh, Kate Baumli, Shariq Iqbal, Colton Bishop, Rebecca Roelofs, Lei~M.
  Zhang, Kay McKinney, Disha Shrivastava, Cosmin Paduraru, George Tucker, Doina
  Precup, Feryal Behbahani, and Aleksandra Faust.
\newblock Training language models to self-correct via reinforcement learning.
\newblock In \emph{International Conference on Learning Representations
  (ICLR)}, 2025.
\newblock URL \url{https://openreview.net/forum?id=CjwERcAU7w}.
\newblock Oral.

\bibitem[Kwon et~al.(2023)Kwon, Li, Zhuang, Sheng, Zheng, Yu, Gonzalez, Zhang,
  and Stoica]{kwon2023efficient}
Woosuk Kwon, Zhuohan Li, Siyuan Zhuang, Ying Sheng, Lianmin Zheng, Cody~Hao Yu,
  Joseph~E. Gonzalez, Hao Zhang, and Ion Stoica.
\newblock Efficient memory management for large language model serving with
  {PagedAttention}.
\newblock In \emph{Proceedings of the 29th Symposium on Operating Systems
  Principles (SOSP)}, pp.\  611--626, 2023.
\newblock \doi{10.1145/3600006.3613165}.

\bibitem[Lambert et~al.(2025)Lambert, Morrison, Pyatkin, Huang, Ivison,
  Brahman, Miranda, Liu, Dziri, Lyu, et~al.]{lambert2025tulu}
Nathan Lambert, Jacob Morrison, Valentina Pyatkin, Shengyi Huang, Hamish
  Ivison, Faeze Brahman, Lester James~V. Miranda, Alisa Liu, Nouha Dziri, Shane
  Lyu, et~al.
\newblock Tulu 3: Pushing frontiers in open language model post-training.
\newblock In \emph{Conference on Language Modeling (COLM)}, 2025.
\newblock URL \url{https://openreview.net/forum?id=i1uGbfHHpH}.

\bibitem[Lightman et~al.(2024)Lightman, Kosaraju, Burda, Edwards, Baker, Lee,
  Leike, Schulman, Sutskever, and Cobbe]{lightman2024lets}
Hunter Lightman, Vineet Kosaraju, Yura Burda, Harri Edwards, Bowen Baker, Teddy
  Lee, Jan Leike, John Schulman, Ilya Sutskever, and Karl Cobbe.
\newblock Let's verify step by step.
\newblock In \emph{International Conference on Learning Representations
  (ICLR)}, 2024.
\newblock URL \url{https://openreview.net/forum?id=v8L0pN6EOi}.

\bibitem[Lin et~al.(2020)Lin, Wohlwend, Chen, and Lei]{lin2020autoregressive}
Alexander Lin, Jeremy Wohlwend, Howard Chen, and Tao Lei.
\newblock Autoregressive knowledge distillation through imitation learning.
\newblock In \emph{Proceedings of the 2020 Conference on Empirical Methods in
  Natural Language Processing (EMNLP)}, pp.\  6121--6133, 2020.
\newblock URL \url{https://aclanthology.org/2020.emnlp-main.494/}.

\bibitem[Lopez-Paz et~al.(2016)Lopez-Paz, Bottou, Sch{\"o}lkopf, and
  Vapnik]{lopezpaz2016unifying}
David Lopez-Paz, L{\'e}on Bottou, Bernhard Sch{\"o}lkopf, and Vladimir Vapnik.
\newblock Unifying distillation and privileged information.
\newblock In \emph{International Conference on Learning Representations
  (ICLR)}, 2016.
\newblock URL \url{https://arxiv.org/abs/1511.03643}.

\bibitem[Loshchilov \& Hutter(2019)Loshchilov and
  Hutter]{loshchilov2019decoupled}
Ilya Loshchilov and Frank Hutter.
\newblock Decoupled weight decay regularization.
\newblock In \emph{International Conference on Learning Representations
  (ICLR)}, 2019.
\newblock URL \url{https://openreview.net/forum?id=Bkg6RiCqY7}.

\bibitem[Lu \& {Thinking Machines Lab}(2025)Lu and {Thinking Machines
  Lab}]{lu2026opd}
Kevin Lu and {Thinking Machines Lab}.
\newblock On-policy distillation.
\newblock Thinking Machines Lab: Connectionism, 2025.
\newblock URL \url{https://thinkingmachines.ai/blog/on-policy-distillation/}.
\newblock Published Oct 27, 2025; doi:10.64434/tml.20251026.

\bibitem[Madaan et~al.(2023)Madaan, Tandon, Gupta, Hallinan, Gao, Wiegreffe,
  Alon, Dziri, Prabhumoye, Yang, Gupta, Majumder, Hermann, Welleck,
  Yazdanbakhsh, and Clark]{madaan2023selfrefine}
Aman Madaan, Niket Tandon, Prakhar Gupta, Skyler Hallinan, Luyu Gao, Sarah
  Wiegreffe, Uri Alon, Nouha Dziri, Shrimai Prabhumoye, Yiming Yang, Shashank
  Gupta, Bodhisattwa~Prasad Majumder, Katherine Hermann, Sean Welleck, Amir
  Yazdanbakhsh, and Peter Clark.
\newblock Self-refine: Iterative refinement with self-feedback.
\newblock In \emph{Advances in Neural Information Processing Systems
  (NeurIPS)}, 2023.
\newblock URL \url{https://arxiv.org/abs/2303.17651}.

\bibitem[Ouyang et~al.(2022)Ouyang, Wu, Jiang, Almeida, Wainwright, Mishkin,
  Zhang, Agarwal, Slama, Ray, et~al.]{ouyang2022training}
Long Ouyang, Jeffrey Wu, Xu~Jiang, Diogo Almeida, Carroll Wainwright, Pamela
  Mishkin, Chong Zhang, Sandhini Agarwal, Katarina Slama, Alex Ray, et~al.
\newblock Training language models to follow instructions with human feedback.
\newblock In \emph{Advances in Neural Information Processing Systems
  (NeurIPS)}, 2022.
\newblock URL
  \url{https://proceedings.neurips.cc/paper_files/paper/2022/hash/b1efde53be364a73914f58805a001731-Abstract-Conference.html}.

\bibitem[Qu et~al.(2024)Qu, Zhang, Garg, and Kumar]{qu2024recursive}
Yuxiao Qu, Tianjun Zhang, Naman Garg, and Aviral Kumar.
\newblock Recursive introspection: Teaching language model agents how to
  self-improve.
\newblock In \emph{Advances in Neural Information Processing Systems
  (NeurIPS)}, 2024.
\newblock URL \url{https://openreview.net/forum?id=DRC9pZwBwR}.

\bibitem[{Qwen Team}(2025)]{yang2025qwen3}
{Qwen Team}.
\newblock Qwen3 technical report.
\newblock \emph{arXiv preprint arXiv:2505.09388}, 2025.
\newblock URL \url{https://arxiv.org/abs/2505.09388}.

\bibitem[Rein et~al.(2024)Rein, Hou, Stickland, Petty, Pang, Dirani, Michael,
  and Bowman]{rein2024gpqa}
David Rein, Betty~Li Hou, Asa~Cooper Stickland, Jackson Petty, Richard~Yuanzhe
  Pang, Julien Dirani, Julian Michael, and Samuel~R. Bowman.
\newblock {GPQA}: A graduate-level google-proof {Q\&A} benchmark.
\newblock In \emph{Conference on Language Modeling (COLM)}, 2024.
\newblock URL \url{https://openreview.net/forum?id=Ti67584b98}.

\bibitem[Ross et~al.(2011)Ross, Gordon, and Bagnell]{ross2011reduction}
St{\'e}phane Ross, Geoffrey~J. Gordon, and Drew Bagnell.
\newblock A reduction of imitation learning and structured prediction to
  no-regret online learning.
\newblock In \emph{Proceedings of the 14th International Conference on
  Artificial Intelligence and Statistics (AISTATS)}, pp.\  627--635, 2011.
\newblock URL \url{https://proceedings.mlr.press/v15/ross11a.html}.

\bibitem[Schulman et~al.(2017)Schulman, Wolski, Dhariwal, Radford, and
  Klimov]{schulman2017proximal}
John Schulman, Filip Wolski, Prafulla Dhariwal, Alec Radford, and Oleg Klimov.
\newblock Proximal policy optimization algorithms.
\newblock \emph{arXiv preprint arXiv:1707.06347}, 2017.
\newblock URL \url{https://arxiv.org/abs/1707.06347}.

\bibitem[Shao et~al.(2024)Shao, Wang, Zhu, Xu, Song, Bi, Zhang, Zhang, Li, Wu,
  and Guo]{shao2024deepseekmath}
Zhihong Shao, Peiyi Wang, Qihao Zhu, Runxin Xu, Junxiao Song, Xiao Bi, Haowei
  Zhang, Mingchuan Zhang, Y.~K. Li, Y.~Wu, and Daya Guo.
\newblock Deepseekmath: Pushing the limits of mathematical reasoning in open
  language models.
\newblock \emph{arXiv preprint arXiv:2402.03300}, 2024.
\newblock URL \url{https://arxiv.org/abs/2402.03300}.

\bibitem[Shinn et~al.(2023)Shinn, Cassano, Berman, Gopinath, Narasimhan, and
  Yao]{shinn2023reflexion}
Noah Shinn, Federico Cassano, Edward Berman, Ashwin Gopinath, Karthik
  Narasimhan, and Shunyu Yao.
\newblock Reflexion: Language agents with verbal reinforcement learning.
\newblock In \emph{Advances in Neural Information Processing Systems
  (NeurIPS)}, 2023.
\newblock URL \url{https://arxiv.org/abs/2303.11366}.

\bibitem[Singh et~al.(2024)Singh, Co-Reyes, Agarwal, Anand, Patil, Garcia, Liu,
  Harrison, Lee, Xu, et~al.]{singh2024beyond}
Avi Singh, John~D. Co-Reyes, Rishabh Agarwal, Ankesh Anand, Piyush Patil,
  Xavier Garcia, Peter~J. Liu, James Harrison, Jaehoon Lee, Kelvin Xu, et~al.
\newblock Beyond human data: Scaling self-training for problem-solving with
  language models.
\newblock \emph{Transactions on Machine Learning Research (TMLR)}, 2024.
\newblock URL \url{https://openreview.net/forum?id=lNAyUngGFK}.

\bibitem[Vapnik \& Vashist(2009)Vapnik and Vashist]{vapnik2009new}
Vladimir Vapnik and Akshay Vashist.
\newblock A new learning paradigm: Learning using privileged information.
\newblock \emph{Neural Networks}, 22\penalty0 (5-6):\penalty0 544--557, 2009.
\newblock \doi{10.1016/j.neunet.2009.06.042}.

\bibitem[von Werra et~al.(2020)von Werra, Belkada, Tunstall, Beeching, Thrush,
  Lambert, Huang, Rasul, and Gallou{\'e}dec]{vonwerra2020trl}
Leandro von Werra, Younes Belkada, Lewis Tunstall, Edward Beeching, Tristan
  Thrush, Nathan Lambert, Shengyi Huang, Kashif Rasul, and Quentin
  Gallou{\'e}dec.
\newblock {TRL}: Transformer reinforcement learning.
\newblock \url{https://github.com/huggingface/trl}, 2020.

\bibitem[Wang et~al.(2026)Wang, Yan, Shen, Liu, Fan, Li, Miao, Yuan, Shen, and
  Ye]{wang2026motab}
Bing Wang, Shaotian Yan, Chen Shen, Kaiyuan Liu, Sinan Fan, Ximing Li, Rui
  Miao, Xiaosong Yuan, Zhanming Shen, and Jieping Ye.
\newblock Backtracking when it strays: Mitigating dual exposure biases in {LLM}
  reasoning distillation.
\newblock \emph{arXiv preprint arXiv:2605.19433}, 2026.
\newblock URL \url{https://arxiv.org/abs/2605.19433}.

\bibitem[Wang et~al.(2024{\natexlab{a}})Wang, Li, Shao, Xu, Dai, Li, Chen, Wu,
  and Sui]{wang2024mathshepherd}
Peiyi Wang, Lei Li, Zhihong Shao, Runxin Xu, Damai Dai, Yifei Li, Deli Chen,
  Yu~Wu, and Zhifang Sui.
\newblock Math-shepherd: Verify and reinforce {LLM}s step-by-step without human
  annotations.
\newblock In \emph{Proceedings of the 62nd Annual Meeting of the Association
  for Computational Linguistics (Volume 1: Long Papers)}, pp.\  9426--9439,
  2024{\natexlab{a}}.
\newblock URL \url{https://aclanthology.org/2024.acl-long.510/}.

\bibitem[Wang et~al.(2024{\natexlab{b}})Wang, Ma, Zhang, Ni, Chandra, Guo, Ren,
  Arulraj, He, Jiang, et~al.]{wang2024mmlupro}
Yubo Wang, Xueguang Ma, Ge~Zhang, Yuansheng Ni, Abhranil Chandra, Shiguang Guo,
  Weiming Ren, Aaran Arulraj, Xuan He, Ziyan Jiang, et~al.
\newblock {MMLU-Pro}: A more robust and challenging multi-task language
  understanding benchmark.
\newblock In \emph{Advances in Neural Information Processing Systems (NeurIPS),
  Datasets and Benchmarks Track}, 2024{\natexlab{b}}.
\newblock URL
  \url{https://proceedings.neurips.cc/paper_files/paper/2024/hash/ad236edc564f3e3156e1b2feafb99a24-Abstract-Datasets_and_Benchmarks_Track.html}.

\bibitem[Wood et~al.(1976)Wood, Bruner, and Ross]{wood1976role}
David Wood, Jerome~S. Bruner, and Gail Ross.
\newblock The role of tutoring in problem solving.
\newblock \emph{Journal of Child Psychology and Psychiatry}, 17\penalty0
  (2):\penalty0 89--100, 1976.
\newblock \doi{10.1111/j.1469-7610.1976.tb00381.x}.

\bibitem[Xia et~al.(2026)Xia, Xu, Yao, McAuley, and He]{xia2026hill}
Yu~Xia, Canwen Xu, Zhewei Yao, Julian McAuley, and Yuxiong He.
\newblock Learning to hint for reinforcement learning.
\newblock \emph{arXiv preprint arXiv:2604.00698}, 2026.
\newblock URL \url{https://arxiv.org/abs/2604.00698}.

\bibitem[Xu et~al.(2026)Xu, Xu, Hong, Ni, Li, Qiu, Lu, and
  Shen]{xu2026relayopd}
Haolei Xu, Xiaowen Xu, Haiwen Hong, Zixuan Ni, Hongxing Li, Yiwen Qiu, Weiming
  Lu, and Yongliang Shen.
\newblock Pass the baton: Trajectory-relayed on-policy distillation.
\newblock \emph{arXiv preprint arXiv:2607.26057}, 2026.
\newblock URL \url{https://arxiv.org/abs/2607.26057}.

\bibitem[Yang et~al.(2025)Yang, Yu, Zhang, Xu, Gonzalez, Cui, and
  Yan]{yang2025supercorrect}
Ling Yang, Zhaochen Yu, Tianjun Zhang, Minkai Xu, Joseph~E. Gonzalez, Bin Cui,
  and Shuicheng Yan.
\newblock Supercorrect: Advancing small {LLM} reasoning with thought template
  distillation and self-correction.
\newblock In \emph{International Conference on Learning Representations
  (ICLR)}, 2025.
\newblock URL \url{https://openreview.net/forum?id=PyjZO7oSw2}.

\bibitem[Yeo et~al.(2026)Yeo, Choi, Ki, and Hwang]{yeo2026hintsd}
Woongyeong Yeo, Yumin Choi, Taekyung Ki, and Sung~Ju Hwang.
\newblock {HINT-SD}: Targeted hindsight self-distillation for long-horizon
  agents.
\newblock In \emph{Findings of the Association for Computational Linguistics:
  EMNLP 2026}, 2026.
\newblock URL \url{https://arxiv.org/abs/2605.17873}.
\newblock arXiv:2605.17873.

\bibitem[Yu et~al.(2025)Yu, Zhang, Zhu, Yuan, Zuo, Yue, Dai, Fan, Liu, Liu,
  Liu, Liu, Lin, Lin, Ma, Sheng, Tong, Zhang, Zhang, Zhang, Zhang, Zhu, Zhu,
  Chen, Chen, Wang, Yu, Song, Wei, Zhou, Liu, Ma, Zhang, Yan, Qiao, Wu, and
  Wang]{yu2025dapo}
Qiying Yu, Zheng Zhang, Ruofei Zhu, Yufeng Yuan, Xiaochen Zuo, Yu~Yue, Weinan
  Dai, Tiantian Fan, Gaohong Liu, Juncai Liu, Lingjun Liu, Xin Liu, Haibin Lin,
  Zhiqi Lin, Bole Ma, Guangming Sheng, Yuxuan Tong, Chi Zhang, Mofan Zhang,
  Ru~Zhang, Wang Zhang, Hang Zhu, Jinhua Zhu, Jiaze Chen, Jiangjie Chen,
  Chengyi Wang, Hongli Yu, Yuxuan Song, Xiangpeng Wei, Hao Zhou, Jingjing Liu,
  Wei-Ying Ma, Ya-Qin Zhang, Lin Yan, Mu~Qiao, Yonghui Wu, and Mingxuan Wang.
\newblock Dapo: An open-source {LLM} reinforcement learning system at scale.
\newblock In \emph{Advances in Neural Information Processing Systems
  (NeurIPS)}, 2025.
\newblock URL \url{https://openreview.net/forum?id=2a36EMSSTp}.

\bibitem[Zelikman et~al.(2022)Zelikman, Wu, Mu, and Goodman]{zelikman2022star}
Eric Zelikman, Yuhuai Wu, Jesse Mu, and Noah~D. Goodman.
\newblock {STaR}: Bootstrapping reasoning with reasoning.
\newblock In \emph{Advances in Neural Information Processing Systems
  (NeurIPS)}, 2022.
\newblock URL
  \url{https://proceedings.neurips.cc/paper_files/paper/2022/hash/639a9a172c044fbb64175b5fad42e9a5-Abstract-Conference.html}.

\bibitem[Zhang et~al.(2026{\natexlab{a}})Zhang, Lv, Li, Wang, Wang, Hu, and
  Yan]{zhang2025stephint}
Kaiyi Zhang, Ang Lv, Jinpeng Li, Yongbo Wang, Feng Wang, Haoyuan Hu, and Rui
  Yan.
\newblock {StepHint}: Multi-level stepwise hints enhance reinforcement learning
  to reason.
\newblock In \emph{Proceedings of the 64th Annual Meeting of the Association
  for Computational Linguistics (Volume 1: Long Papers)}, pp.\  37846--37864,
  2026{\natexlab{a}}.
\newblock \doi{10.18653/v1/2026.acl-long.1755}.
\newblock URL \url{https://aclanthology.org/2026.acl-long.1755/}.

\bibitem[Zhang et~al.(2026{\natexlab{b}})Zhang, Wu, Zhu, Tan, Yu, He, and
  Jia]{zhang2026scafgrpo}
Xichen Zhang, Sitong Wu, Yinghao Zhu, Haoru Tan, Shaozuo Yu, Ziyi He, and Jiaya
  Jia.
\newblock {Scaf-GRPO}: Scaffolded group relative policy optimization for
  enhancing {LLM} reasoning.
\newblock In \emph{International Conference on Learning Representations
  (ICLR)}, 2026{\natexlab{b}}.
\newblock URL \url{https://openreview.net/forum?id=bOwVr0yr7r}.

\bibitem[Zhang et~al.(2026{\natexlab{c}})Zhang, Lin, and Wu]{zhang2026stepopsd}
Yanfei Zhang, Xu~Lin, and Chenglin Wu.
\newblock {StepOPSD}: Step-aware online preference self-distillation for agent
  reinforcement learning.
\newblock In \emph{Findings of the Association for Computational Linguistics:
  EMNLP 2026}, 2026{\natexlab{c}}.
\newblock URL \url{https://arxiv.org/abs/2605.27140}.
\newblock arXiv:2605.27140.

\bibitem[Zhao et~al.(2026)Zhao, Xie, Liu, Huang, Pang, Chen, and
  Grover]{zhao2026selfdistilled}
Siyan Zhao, Zhihui Xie, Mengchen Liu, Jing Huang, Guan Pang, Feiyu Chen, and
  Aditya Grover.
\newblock Self-distilled reasoner: On-policy self-distillation for large
  language models.
\newblock \emph{arXiv preprint arXiv:2601.18734}, 2026.
\newblock URL \url{https://arxiv.org/abs/2601.18734}.

\bibitem[Zheng et~al.(2026)Zheng, Ma, Liang, Ruan, Fu, Lin, Zhu, Zeng, and
  Cai]{zheng2026scope}
Binbin Zheng, Xing Ma, Yiheng Liang, Jingqing Ruan, Xiaoliang Fu, Kepeng Lin,
  Benchang Zhu, Ke~Zeng, and Xunliang Cai.
\newblock Scope: Signal-calibrated on-policy distillation enhancement with
  dual-path adaptive weighting.
\newblock \emph{arXiv preprint arXiv:2604.10688}, 2026.
\newblock URL \url{https://arxiv.org/abs/2604.10688}.

\end{thebibliography}
\bibliographystyle{iclr2027_conference}

\appendix
\section{Complete Training Algorithm}
\label{sec:appendix-algorithm}

Algorithm~\ref{alg:full} gives the complete procedure that \methodname{} runs
on one training minibatch, from the initial on-policy rollouts to the returned
training loss. The method definitions and design rationale remain in
Section~\ref{sec:method}.

\begin{algorithm}[H]
\caption{\methodname{}: multi-round repair and stage-weighted distillation}
\label{alg:full}
\begin{algorithmic}[1]
\STATE \textbf{Input:} minibatch $\mathcal B$ of problems $q$ with reference
solutions $z$, student $p_\theta$, frozen teacher $p_{\mathrm T}$, outcome verifier
$V$, and guidance generator $\mathcal G$
\STATE \textbf{Hyperparameters:} repair budget $K$, endpoint tolerance
$\epsilon\in(0,1)$, temporal factor $\gamma\in(0,1]$, and outcome
bonus $\rho\geq 1$
\STATE \textbf{Output:} minibatch training loss
$\mathcal L_{\methodname}$
\FORALL{$(q,z)\in\mathcal B$, indexed by $n$}
  \STATE $\mathcal S_n\gets\varnothing$, $\mathcal L_n\gets 0$, $t\gets 0$,
  $\mathcal H_0\gets\operatorname{Render}(q)$, and sample
  $y_0\sim p_\theta(\cdot\mid\mathcal H_0)$
  \WHILE{$t<K$ and $y_t$ is a well-formed response with $V(y_t)=0$}
    \STATE $(a_t,g_t)\gets\mathcal G(q,z,\mathcal H_t,y_t)$
    \IF{$(a_t,g_t)$ fails validation}
      \STATE \textbf{break}
    \ENDIF
    \STATE $b_t\gets$ the exact match position of $a_t$ in $y_t$
    \STATE $e_t\gets$ the earliest block endpoint with corrective coverage
    $1-\epsilon$
    \STATE Compute $\mathcal L_{n,t}$ on the repair region
    $R_t=y_t[b_t:e_t]$ and add $(t,\mathcal L_{n,t})$ to $\mathcal S_n$
    \STATE $\mathcal H_{t+1}\gets
    \operatorname{Update}(\mathcal H_t,y_t,g_t)$ and, if $t<K-1$, sample
    $y_{t+1}\sim p_\theta(\cdot\mid\mathcal H_{t+1})$
    \STATE $s_{n,t}\gets\mathds{1}[V(y_{t+1})=1]$, or $0$ if no retry is
    sampled, and $t\gets t+1$
  \ENDWHILE
  \IF{$\mathcal S_n\neq\varnothing$}
    \STATE $T_n\gets t-1$
    \STATE $w_{n,t}\gets\gamma^{t}\rho^{s_{n,t}}
    \big/\sum_{k=0}^{T_n}\gamma^{k}\rho^{s_{n,k}}$ for $t=0,\ldots,T_n$
    \STATE $\mathcal L_n\gets\sum_{t=0}^{T_n}w_{n,t}\mathcal L_{n,t}$
  \ENDIF
\ENDFOR
\RETURN $\mathcal L_{\methodname}
=\frac{1}{|\mathcal B|}\sum_{n\in\mathcal B}\mathcal L_n$
\end{algorithmic}
\end{algorithm}

\section{Prompts and Validation Details}
\label{sec:appendix-prompts}

\subsection{Guidance generator instruction}

\begin{figure}[!t]
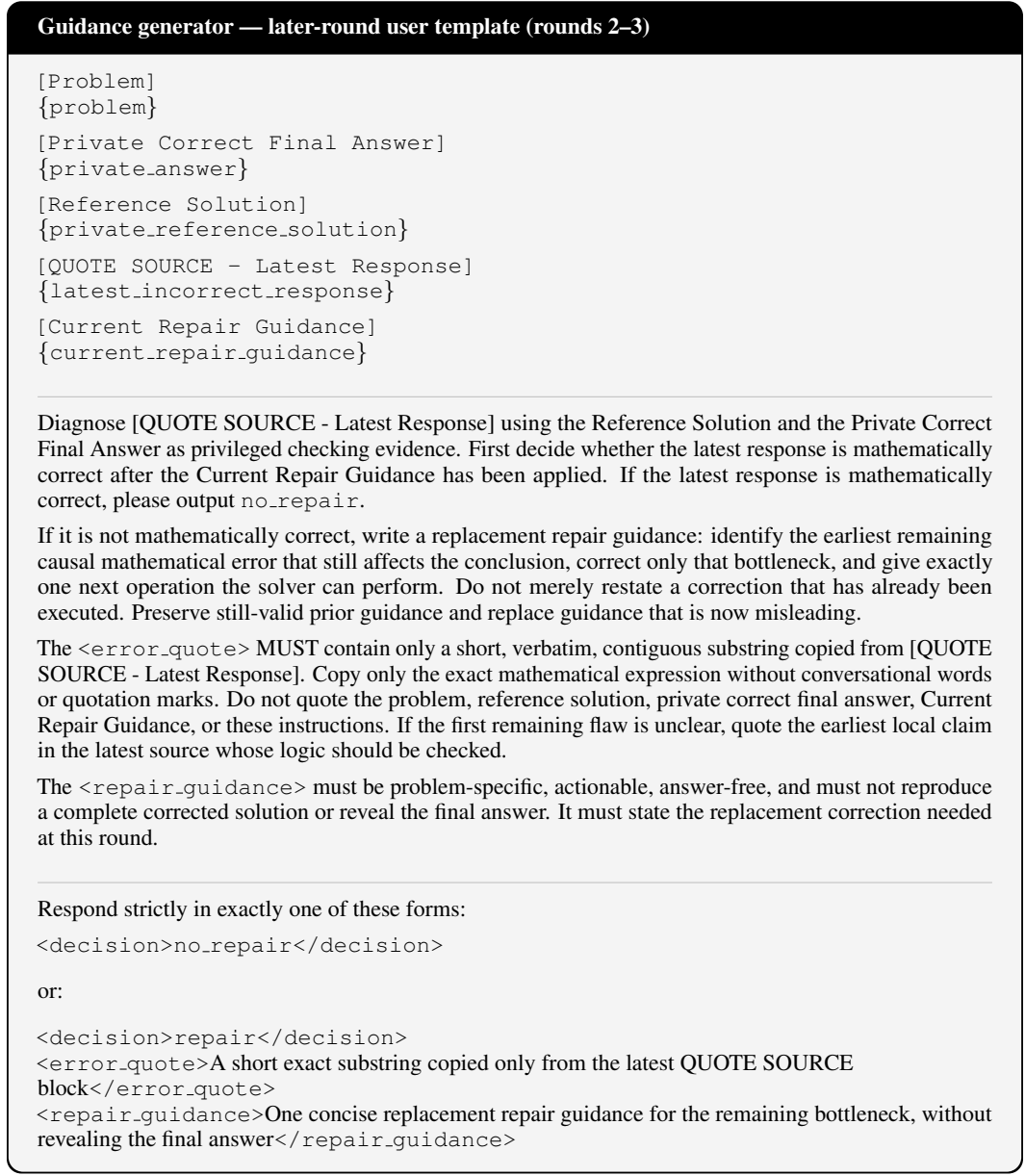

\centering
\begin{promptbox}[unbreakable]{Guidance generator --- later-round user template (rounds 2--3)}
\texttt{[Problem]}\\
\texttt{\{problem\}}\\[4pt]
\texttt{[Private Correct Final Answer]}\\
\texttt{\{private\_answer\}}\\[4pt]
\texttt{[Reference Solution]}\\
\texttt{\{private\_reference\_solution\}}\\[4pt]
\texttt{[QUOTE SOURCE - Latest Response]}\\
\texttt{\{latest\_incorrect\_response\}}\\[4pt]
\texttt{[Current Repair Guidance]}\\
\texttt{\{current\_repair\_guidance\}}

\promptrule

Diagnose [QUOTE SOURCE - Latest Response] using the Reference Solution and
the Private Correct Final Answer as privileged checking evidence. First
decide whether the latest response is mathematically correct after the
Current Repair Guidance has been applied. If the latest response is
mathematically correct, please output \texttt{no\_repair}.

If it is not
mathematically correct, write a replacement repair guidance: identify the
earliest remaining causal mathematical error that still affects the
conclusion, correct only that bottleneck, and give exactly one next operation
the solver can perform. Do not merely restate a correction that has already
been executed. Preserve still-valid prior guidance and replace guidance that
is now misleading.

The \texttt{<error\_quote>} MUST contain only a short, verbatim, contiguous
substring copied from [QUOTE SOURCE - Latest Response]. Copy only the exact
mathematical expression without conversational words or quotation marks. Do
not quote the problem, reference solution, private correct final answer,
Current Repair Guidance, or these instructions. If the first remaining flaw is unclear,
quote the earliest local claim in the latest source whose logic should be
checked.

The \texttt{<repair\_guidance>} must be problem-specific, actionable,
answer-free, and must not reproduce a complete corrected solution or reveal
the final answer. It must state the replacement correction needed at this
round.

\promptrule

Respond strictly in exactly one of these forms:

{\raggedright
\texttt{<decision>no\_repair</decision>}\par
\vspace{4pt}
or:\par
\vspace{4pt}
\texttt{<decision>repair</decision>}\\
\texttt{<error\_quote>}A short exact substring copied only from the latest
QUOTE SOURCE block\texttt{</error\_quote>}\\
\texttt{<repair\_guidance>}One concise replacement repair guidance for the
remaining bottleneck, without revealing the final
answer\texttt{</repair\_guidance>}\par}
\end{promptbox}
\caption{\textbf{Guidance generator, later repair rounds.} The
template replaces the current guidance after a retry that is still
incorrect, and additionally receives the guidance the student has already
seen.}
\label{fig:prompt-generator-later}
\end{figure}

Figures~\ref{fig:prompt-generator-first} and~\ref{fig:prompt-generator-later}
show the full user templates of the guidance generator for the first
repair round and for the later rounds. At every round the generator receives
the problem, the reference solution and its final answer as privileged
checking evidence, and the latest failed response. From the second round it
also receives the guidance the student has already seen. Braced fields are
filled at runtime.

\begin{figure}[!t]
\centering
\begin{promptbox}[unbreakable]{Guidance generator --- first-round user template}
\texttt{[Problem]}\\
\texttt{\{problem\}}\\[4pt]
\texttt{[Private Correct Final Answer]}\\
\texttt{\{private\_answer\}}\\[4pt]
\texttt{[Reference Solution]}\\
\texttt{\{private\_reference\_solution\}}\\[4pt]
\texttt{[QUOTE SOURCE - Incorrect Solution]}\\
\texttt{\{incorrect\_response\}}

\promptrule

Diagnose [QUOTE SOURCE - Incorrect Solution] using the Reference Solution and
the Private Correct Final Answer as privileged checking evidence. First
decide whether the quoted response is mathematically correct.

If the response is not mathematically correct, identify the earliest causal
mathematical error that still affects the conclusion. The
\texttt{<error\_quote>} MUST contain only a short, verbatim, contiguous
substring copied from [QUOTE SOURCE - Incorrect Solution]. Copy only the
exact mathematical expression without conversational words or quotation
marks. Do not quote the problem, reference solution, private correct final
answer, or these instructions. If the first flawed step is unclear, quote the earliest local
claim in the source whose logic should be checked.

The \texttt{<repair\_guidance>} must identify the exact mistake, explain how
to fix it, and state the correct governing logic. It must be
problem-specific, actionable, answer-free, and must not reproduce a complete
corrected solution or reveal the final answer.

When the reasoning is mathematically correct, please output
\texttt{no\_repair}.

\promptrule

Respond strictly in exactly one of these forms:

{\raggedright
\texttt{<decision>no\_repair</decision>}\par
\vspace{4pt}
or:\par
\vspace{4pt}
\texttt{<decision>repair</decision>}\\
\texttt{<error\_quote>}A short exact substring copied only from the QUOTE
SOURCE block\texttt{</error\_quote>}\\
\texttt{<repair\_guidance>}Identify the exact mistake in the quote, explain
how to fix it, and state the correct logic without revealing the final
answer\texttt{</repair\_guidance>}\par}
\end{promptbox}
\caption{\textbf{Guidance generator, first repair round.} The
template diagnoses the initial failed response $y_0$ under the privileged
reference solution and final answer.}
\label{fig:prompt-generator-first}
\end{figure}

\subsection{Repair dialogue}

The retry is sampled from a multi-turn dialogue whose turns are shown below.
The initial user turn poses the problem, and each unsuccessful response is
followed by a feedback turn that delivers the current repair guidance. From
the second feedback turn onward the opening line instead reads
\texttt{The previous answer is still incorrect.}

\begin{promptbox}{Repair dialogue --- initial problem turn}
Problem: \texttt{\{problem\}}\\
Please reason step by step, and put your final answer within
\texttt{\textbackslash boxed\{\}}.
\end{promptbox}

\begin{promptbox}[unbreakable]{Repair dialogue --- feedback turn after an unsuccessful response}
The previous answer is incorrect.

Use the following repair guidance, which explains how to correct the reasoning:\\
\texttt{=== Repair Guidance Begin ===}\\
\texttt{\{current\_repair\_guidance\}}\\
\texttt{=== Repair Guidance End ===}

Revise the reasoning and answer again. Do not simply repeat the previous
attempt.

Please reason step by step, and put your final answer within
\texttt{\textbackslash boxed\{\}}.
\end{promptbox}

\subsection{Validation rules}

The generator output is accepted only when it matches one of the two
declared forms in full, with a nonempty error quote and a nonempty
repair-guidance field. The error quote must be located in the latest
response, and guidance that states the final answer is rejected. The leak
check excludes the error quote and rejects guidance containing boxed
answers or rule-based matches to the normalized reference answer, such
as explicit answer statements or variable assignments. These surface-form
checks do not guarantee detection of all semantic answer leakage. When any
check fails, or when the generator returns \texttt{no\_repair}, the round
yields no supervised stage and stops further retries.

\section{Training and Evaluation Details}
\label{sec:appendix-evaluation}

\subsection{Training settings}
\label{sec:appendix-training}

Following the comparison setup of OPSD \citep{zhao2026selfdistilled},
Table~\ref{tab:training-comparison} places SFT, GRPO, OPSD, and \methodname{}
side by side, grouping their optimization and on-policy sampling settings.
All trained baselines in Table~\ref{tab:main-results} are our own
reproductions on the same training corpus and backbones as \methodname{};
the GRPO reproduction applies no KL penalty, and the untrained base-model
row has no optimization settings.
Training problems come from DAPO-Math-17K \citep{yu2025dapo}, and outcome
verification at every repair round uses rule-based answer matching.
Reference solutions are generated by GPT-5.5 and retained
only when their final answers verify against the dataset answers, leaving
15{,}650 of the 17{,}398 problems as training rows. SFT minimizes
cross-entropy on these reference solutions without on-policy sampling and
uses a 16{,}000-token maximum sequence length measured over the full
supervised sequence.

All trained methods are implemented on top of TRL \citep{vonwerra2020trl}
and use AdamW, BF16, and FlashAttention-2 \citep{dao2024flashattention},
with a colocated vLLM sampler \citep{kwon2023efficient} for on-policy
rollouts. For OPSD and \methodname{}, the pointwise divergence-clipping
threshold is $\tau=0.05$.

\begin{table}[!t]
\centering
\caption{\textbf{Training settings for SFT, GRPO, OPSD, and \methodname{}.}
Lengths are measured in tokens; the SFT limit applies to the full supervised
sequence. A dash denotes a setting that
is not applicable. For \methodname{}, one rollout is sampled per dialogue
state: the initial response can be followed by at most $K-1$ retries.}
\label{tab:training-comparison}
\small
\setlength{\tabcolsep}{4pt}
\renewcommand{\arraystretch}{1.10}
\begin{tabularx}{\linewidth}{@{}Lcccc@{}}
\toprule
\textbf{Parameter} & \textbf{SFT} & \textbf{GRPO} & \textbf{OPSD} & \textbf{\methodname{}} \\
\midrule
\multicolumn{5}{@{}l}{\textit{Optimization and adaptation}} \\
\addlinespace[2pt]
Optimizer & AdamW & AdamW & AdamW & AdamW \\
Learning rate & $5\times10^{-6}$ & $5\times10^{-6}$ & $5\times10^{-6}$ & $5\times10^{-6}$ \\
Effective batch size & 32 & 32 & 32 & 32 \\
Training precision & BF16 & BF16 & BF16 & BF16 \\
LoRA rank / alpha & 64 / 128 & 64 / 128 & 64 / 128 & 64 / 128 \\
Optimizer steps & 100 & 500 & 100 & 100 \\
SFT maximum sequence length & 16,000 & -- & -- & -- \\
\midrule
\multicolumn{5}{@{}l}{\textit{On-policy sampling}} \\
\addlinespace[2pt]
Maximum completion length & -- & 16,000 & 8,192 & 8,192 \\
Rollouts per prompt / dialogue state & -- & 8 & 1 & 1 \\
Sampling temperature & -- & 1.2 & 1.1 & 1.1 \\
\bottomrule
\end{tabularx}
\end{table}

\paragraph{\methodname{}-specific settings.}
\methodname{} adds a repair loop and a guidance generator on top of the
shared configuration in Table~\ref{tab:training-comparison}. Each problem
allows at most $K=3$ repair rounds, with stage weighting
$(\gamma,\rho)=(0.8,1.5)$ and endpoint tolerance $\epsilon=0.1$. The
student decodes with top-$p$ 0.95 and top-$k$ 20, generating at most
8{,}192 completion tokens under a total sequence-length limit of
19{,}168 tokens. Self-guidance from the current student policy and external
guidance from Qwen3.5-27B share the same inference settings. Both
generate with thinking mode enabled, a temperature of 0.7, top-$p$ 0.9,
and at most 8{,}192 new tokens.

\subsection{Inference settings}
\label{sec:appendix-inference}

Table~\ref{tab:inference-config} summarizes the inference settings shared
by the base model and all trained methods for mathematical-reasoning
evaluation. We use the fixed random seed \texttt{2026} to generate 12
responses per problem, with a sampling temperature of 1.0, top-$p$ 0.95,
no top-$k$ restriction, a maximum of 38,912 new tokens, a model context
length of 40,960, and thinking enabled. Following the OPSD evaluation
protocol \citep{zhao2026selfdistilled}, we evaluate checkpoints every 20
steps up to 100 steps. Overall Math Avg. is the
macro-average of the three mathematical benchmark scores.

\begin{table}[!htbp]
\centering
\caption{\textbf{Inference settings for mathematical-reasoning evaluation.}
The base model and all trained methods use these settings on AIME24,
AIME25, and HMMT25, without repair guidance.}
\label{tab:inference-config}
\small
\setlength{\tabcolsep}{6pt}
\renewcommand{\arraystretch}{1.10}
\begin{tabularx}{0.72\linewidth}{@{}Lc@{}}
\toprule
\textbf{Parameter} & \textbf{Value} \\
\midrule
Maximum new tokens & 38,912 \\
Maximum model length (tokens) & 40,960 \\
Thinking mode & Enabled \\
\midrule
Sampling temperature & 1.0 \\
Top-$p$ & 0.95 \\
Top-$k$ & -- \\
Samples per problem & 12 \\
Random seed & \texttt{2026} \\
\bottomrule
\end{tabularx}
\end{table}

\paragraph{Out-of-distribution evaluation.}
MMLU-Pro is evaluated on a fixed 1{,}000-question subset that covers all
14 subject categories, and GPQA on the 198-question GPQA-Diamond split. For the base model and every trained
method, both benchmarks draw four samples per question (Avg@4) with a
sampling temperature of 1.0, top-$p$ 0.95, no top-$k$ restriction, and the
random seed \texttt{2026071201}. MMLU-Pro applies the chat template with
thinking enabled, a maximum of 8{,}192 new tokens, and a model context
length of 12{,}288. GPQA feeds the question prompt to the model without the
chat template, with a maximum of 38{,}912 new tokens and a model context
length of 40{,}960. The predicted option is read from the last
\texttt{\textbackslash boxed\{\}} in the response, and MMLU-Pro
additionally accepts an explicit final-answer statement.

\begingroup
\small
\section{Additional Results and Examples}
\label{sec:appendix-results}

\subsection{Worked repair examples}
\label{sec:appendix-case}

We reconstruct two complete repair trajectories from the training-time repair
logs of the Qwen3-4B self-guidance run, both repaired at the second round.
In the first the retry fixes the anchored slip but returns the identical
wrong answer, and in the second the retry adopts the corrected equation
yet slips again in the very next step although the guidance is valid. The two trajectories illustrate the rationale of the outcome-aware
stage weighting in Section~\ref{sec:method-objective}. Because a retry
can fail even under valid guidance, a failed retry is not penalized, and
the weighting grants its outcome bonus only to the stage whose guidance
leads to the verified repair. In each round box, the error anchor
$a_t$ located by the guidance generator is
\colorbox{anchorhl}{highlighted} in the failed response, and the supervised
repair region $R_t$ opens at the anchor and ends at the endpoint
Eq.~\ref{eq:endpoint} selects for that stage. Elisions are marked
$[\ldots]$.

\paragraph{Example 1: repaired in two rounds.}
The initial attempt counts $\binom{20}{2}=190$ choices for Terry and
$\binom{18}{2}=153$ for Mary but writes the product as $29{,}170$, counts
only the two-red and two-blue cases, and reduces the resulting fraction
with a faulty remainder chain to $12/139$, answering $151$. The
first-round guidance anchors the earliest error, the multiplication slip;
the retry corrects the product to $29{,}070$ yet repeats the earlier
remainder chain during the reduction and returns the identical $151$, so
$s_0=0$. The second-round guidance targets the remaining omission, the
mixed one-red-one-blue case, and the following retry sums all three cases
and reaches the verified $441$, so $s_1=1$.

\begin{promptbox}{Example 1, round 0 --- failed attempt $y_0$ (verified incorrect)}
\textbf{Problem.} A jar has $10$ red candies and $10$ blue candies. Terry
picks two candies at random, then Mary picks two of the remaining candies
at random. Given that the probability that they get the same color
combination, irrespective of order, is $m/n$, where $m$ and $n$ are
relatively prime positive integers, find $m+n$.

\medskip
$[\ldots]$ (Terry has $\binom{20}{2} = 190$ choices and Mary has
$\binom{18}{2} = 153$)
So the total number of possible pairs of choices (Terry picks first, then
Mary) is:
\[\colorbox{anchorhl}{$190 \times 153 = 29170$}\]
$[\ldots]$ (only two favorable cases are considered, both picking two reds
and both picking two blues, each counted as
$\binom{10}{2}\binom{8}{2} = 45 \times 28 = 1260$)

Total favorable outcomes:
\[1260 + 1260 = 2520\]
$[\ldots]$ (a Euclidean reduction of $2520/29170$ with a faulty remainder
chain gives)
\[\frac{2520}{29170} = \frac{12}{139}\]
So $m = 12$, $n = 139$, and:
\[m + n = \boxed{151}\]
\end{promptbox}

\begin{promptbox}{Example 1 --- generator output after round 0: $a_0$ and $g_0$}
\textbf{Error anchor $a_0$.} \texttt{190 \textbackslash times 153 = 29170}

\medskip
\textbf{Repair guidance $g_0$.} The incorrect solution miscalculates the
total number of possible outcomes as $190 \times 153 = 29170$, but the
correct value is $190 \times 153 = 29070$. This error arises from an
arithmetic mistake in multiplication. $[\ldots]$ The correct logic is to
compute $\binom{20}{2} \times \binom{18}{2} = 190 \times 153 = 29070$,
ensuring the denominator is accurate for the probability computation.
\end{promptbox}

\begin{promptbox}{Example 1, round 1 --- retry $y_1$ (verified incorrect, $s_0=0$)}
$[\ldots]$ (the retry adopts the correction)
So the total number of possible outcomes (Terry's pick $\times$ Mary's
pick) is:
\[190 \times 153 = \boxed{29070}\]
So the correct total number of outcomes is 29070, not 29170.
$[\ldots]$ (the same two cases are counted again)

\colorbox{anchorhl}{Total favorable outcomes}
\[1260 + 1260 = 2520\]
$[\ldots]$ (the reduction repeats the earlier faulty remainder chain,
$29070 \div 2520 = 11$ remainder $1050$, and lands on the same fraction)
\[\frac{2520}{29070} = \frac{2520 \div 210}{29070 \div 210} = \frac{12}{139}\]
Thus,
\[m = 12, \quad n = 139 \Rightarrow m + n = \boxed{151}\]
\end{promptbox}

\begin{promptbox}{Example 1 --- generator output after round 1: $a_1$ and $g_1$}
\textbf{Error anchor $a_1$.} \texttt{Total favorable outcomes}

\medskip
\textbf{Repair guidance $g_1$.} Add the 8100 favorable outcomes from Case
3 (both picking one red and one blue) to the total, resulting in
$1260 + 1260 + 8100 = 10620$
\end{promptbox}

\begin{promptbox}{Example 1, round 2 --- retry $y_2$ (verified correct, $s_1=1$)}
$[\ldots]$ (the retry keeps the corrected total $29070$, recounts Cases 1
and 2, and adds the new case)

\textbf{Case 3: Both pick one red and one blue.}
Terry can pick one red and one blue in
$\binom{10}{1} \times \binom{10}{1} = 100$ ways $[\ldots]$ after Terry's
pick, $9$ reds and $9$ blues remain, so Mary has
$\binom{9}{1} \times \binom{9}{1} = 81$ ways, and the total for this case
is $100 \times 81 = 8100$.

Total favorable outcomes:
\[1260 + 1260 + 8100 = \boxed{10620}\]
$[\ldots]$ (a fresh Euclidean run gives $\gcd = 90$, and $118$ and $323$
are checked to be coprime)
\[\frac{10620}{29070} = \frac{10620 \div 90}{29070 \div 90}
= \frac{118}{323}\]
Hence:
\[m = 118, \quad n = 323 \Rightarrow m + n = \boxed{441}\]
\end{promptbox}

\paragraph{Example 2: a retry that fails under valid guidance.}
The initial attempt equates the binomial probabilities of exactly two and
exactly three heads but mishandles the $(3/4)$ exponent while cancelling
the common factor, reducing the equality to
$\binom{n}{2}=\binom{n}{3}\cdot\frac{3}{16}$ and reaching $n=18$. The
first-round guidance anchors this equation and derives the correct
coefficient $\frac{1}{3}$. The retry derives the correct equation but
mishandles the right-hand side while clearing denominators. The next
guidance recommends multiplying by 18 to clear all denominators in one
step, after which the retry succeeds.

\begin{promptbox}{Example 2, round 0 --- failed attempt $y_0$ (verified incorrect)}
\textbf{Problem.} An unfair coin lands on heads with a probability of
$\tfrac{1}{4}$. When tossed $n$ times, the probability of exactly two
heads is the same as the probability of exactly three heads. What is the
value of $n$?

\medskip
$[\ldots]$ (the response equates the two binomial probabilities)
\[\binom{n}{2} \left(\tfrac{1}{4}\right)^2 \left(\tfrac{3}{4}\right)^{n-2}
= \binom{n}{3} \left(\tfrac{1}{4}\right)^3 \left(\tfrac{3}{4}\right)^{n-3}\]
$[\ldots]$ (dividing both sides by the common factor
$(\tfrac{1}{4})^2 (\tfrac{3}{4})^{n-2}$, the equation becomes)
\[\binom{n}{2} = \binom{n}{3} \cdot \frac{1}{4} \cdot \frac{3}{4}\]
So now, simplify:
\[\colorbox{anchorhl}{$\binom{n}{2} = \binom{n}{3} \cdot \frac{3}{16}$}\]
Now plug them in:
\[\frac{n(n-1)}{2} = \frac{n(n-1)(n-2)}{6} \cdot \frac{3}{16}\]
$[\ldots]$ (multiplying by $6$ and dividing by $3n(n-1)$ gives)
\[1 = \frac{1}{16}(n-2) \Rightarrow 16 = n - 2 \Rightarrow n = 18\]
\textbf{Final Answer:}
\[\boxed{18}\]
\end{promptbox}

\begin{promptbox}{Example 2 --- generator output after round 0: $a_0$ and $g_0$}
\textbf{Error anchor $a_0$.} \texttt{binom(n,2) = binom(n,3) *
\textbackslash frac\{3\}\{16\}}

\medskip
\textbf{Repair guidance $g_0$.} The incorrect solution incorrectly
simplifies the exponent terms after dividing by the common factors. The
correct coefficient after simplification should be $\frac{1}{3}$, not
$\frac{3}{16}$. This arises from miscalculating the exponents: the
right-hand side should be $\binom{n}{3} \cdot \frac{1}{4} \cdot
\frac{4}{3}$ (from $(1/4)^1 \cdot (3/4)^{-1}$), which simplifies to
$\frac{1}{3}$. $[\ldots]$ The correct equation after dividing by the
common factor is $\binom{n}{2} = \binom{n}{3} \cdot \frac{1}{3}$, not
$\frac{3}{16}$.
\end{promptbox}

\begin{promptbox}{Example 2, round 1 --- retry $y_1$ (verified incorrect, $s_0=0$)}
$[\ldots]$ (the retry rederives the cancellation with the correct
exponents)
\[\binom{n}{2} = \binom{n}{3} \cdot \left(\tfrac{1}{4}\right)^{3-2}
\cdot \left(\tfrac{3}{4}\right)^{(n-3)-(n-2)}
= \binom{n}{3} \cdot \frac{1}{4} \cdot \frac{4}{3}
= \binom{n}{3} \cdot \frac{1}{3}\]
This is the correct simplified equation.
$[\ldots]$
\[\frac{n(n-1)}{2} = \frac{n(n-1)(n-2)}{6} \cdot \frac{1}{3}\]
\colorbox{anchorhl}{Multiply both sides by 6 to eliminate denominators:}
\[3n(n-1) = \frac{1}{3} n(n-1)(n-2) \cdot 6
\Rightarrow 3n(n-1) = 2n(n-1)(n-2)\]
Divide both sides by $n(n-1)$ $[\ldots]$
\[3 = 2(n-2) \Rightarrow 3 = 2n - 4 \Rightarrow 2n = 7
\Rightarrow n = \boxed{7}\]
\textbf{Final Answer:}
\[\boxed{7}\]
\end{promptbox}

\begin{promptbox}{Example 2 --- generator output after round 1: $a_1$ and $g_1$}
\textbf{Error anchor $a_1$.} \texttt{Multiply both sides by 6 to eliminate
denominators}

\medskip
\textbf{Repair guidance $g_1$.} Correct the multiplication factor to 18
instead of 6 when eliminating denominators in the equation
\end{promptbox}

\begin{promptbox}{Example 2, round 2 --- retry $y_2$ (verified correct, $s_1=1$)}
$[\ldots]$ (the retry again reaches
$\binom{n}{2} = \frac{1}{3}\binom{n}{3}$ and substitutes the binomial
expressions)
\[\frac{n(n-1)}{2} = \frac{1}{3} \cdot \frac{n(n-1)(n-2)}{6}
= \frac{n(n-1)(n-2)}{18}\]
Now eliminate $n(n-1)$ from both sides $[\ldots]$
\[\frac{1}{2} = \frac{n-2}{18}\]
Multiply both sides by 18:
\[9 = n - 2 \Rightarrow n = \boxed{11}\]
\textbf{Final Answer:}
\[\boxed{11}\]
\end{promptbox}

\endgroup

\end{document}